\documentclass[10pt,twocolumn,letterpaper]{article}

\usepackage[preprint]{cvpr}      

\usepackage{graphicx}
\usepackage{amsmath}
\usepackage{amssymb}
\usepackage{booktabs}
\usepackage{bm,bbm}
\usepackage[makeroom]{cancel}
\usepackage[table,xcdraw,dvipsnames]{xcolor}
\usepackage{adjustbox}

\newcommand{\settitle}{\@maketitle}

\newcommand{\ourMethod}[0]{PROB}

\usepackage[pagebackref,breaklinks,colorlinks]{hyperref}

\usepackage[capitalize]{cleveref}
\crefname{section}{Sec.}{Secs.}
\Crefname{section}{Section}{Sections}
\Crefname{table}{Table}{Tables}
\crefname{table}{Tab.}{Tabs.}

\def\cvprPaperID{8607} 
\def\confName{CVPR}
\def\confYear{2022}

\begin{document}

\title{\ourMethod: Probabilistic Objectness for Open World Object Detection 
}

\author{Orr Zohar\\
Stanford\\
{\tt\small orrzohar@stanford.edu}
\and
Kuan-Chieh Wang\\
Stanford\\
{\tt\small wangkua1@stanford.edu}
\and
Serena Yeung\\
Stanford\\
{\tt\small syyeung@stanford.edu}
}
\maketitle


\begin{abstract}
Open World Object Detection (OWOD) is a new and challenging computer vision task that bridges the gap between classic object detection (OD) benchmarks and object detection in the real world.
In addition to detecting and classifying \emph{seen/labeled} objects, OWOD algorithms are expected to detect \emph{novel/unknown} objects - which can be classified and incrementally learned.
In standard OD, object proposals not overlapping with a labeled object are automatically classified as background. Therefore, simply applying OD methods to OWOD fails as unknown objects would be predicted as background. 
The challenge of detecting unknown objects stems from the lack of supervision in distinguishing unknown objects and background object proposals. Previous OWOD methods have attempted to overcome this issue by generating supervision using pseudo-labeling - however, unknown object detection has remained low.
Probabilistic/generative models may provide a solution for this challenge. 
Herein, we introduce a novel probabilistic framework for objectness estimation, where we alternate between probability distribution estimation and objectness likelihood maximization of known objects in the embedded feature space - ultimately allowing us to estimate the objectness probability of different proposals. 
The resulting \textbf{Pr}obabilistic \textbf{Ob}jectness transformer-based open-world detector, \ourMethod, integrates our framework into traditional object detection models, adapting them for the open-world setting.
Comprehensive experiments on OWOD benchmarks show that \ourMethod\ outperforms all existing OWOD methods in both unknown object detection ($\sim 2\times$ unknown recall) and known object detection ($\sim 10\%$ mAP). Our code is available at \href{https://github.com/orrzohar/PROB}{https://github.com/orrzohar/PROB}.
\end{abstract}


\section{Introduction}
\label{sec:introduction}
\begin{figure}
    \centering
    \includegraphics[width=0.97\columnwidth]{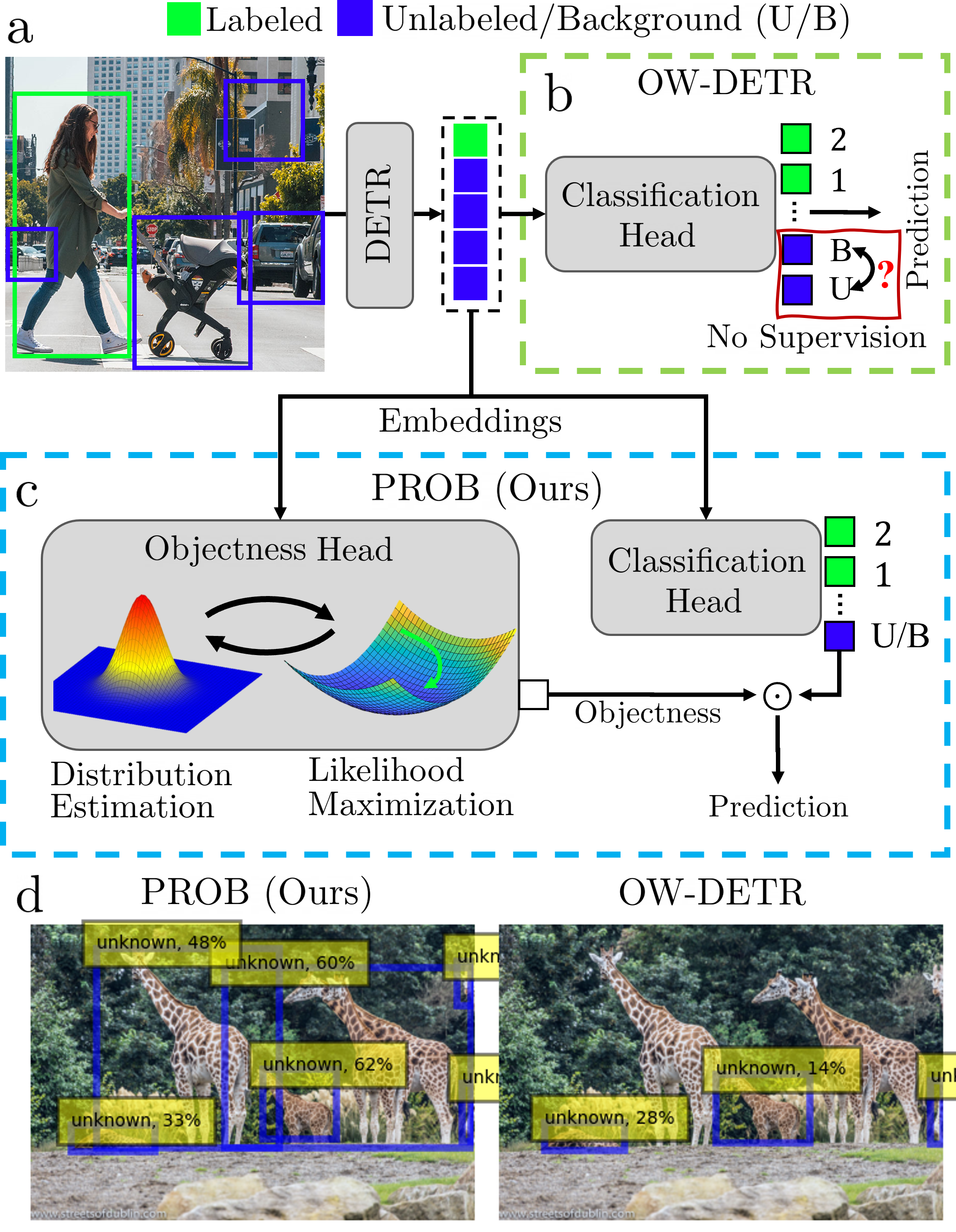}\vspace{-0.2cm}
    \caption{Comparison of \ourMethod\ with other open world object detectors. 
    (a) Query embeddings (each representing a single prediction) are extracted from an image via the deformable DETR model. 
    (b) other open-world detectors attempt to directly distinguish between unlabeled `hidden' objects and background without supervision (red). 
    (c) \ourMethod's scheme of probabilistic objectness training and revised inference, which performs alternation optimization of
    (i) Embeddings distribution estimation and (ii) maximizing the likelihood of embeddings that represent known objects.  
    (d) Qualitative example of the improved unknown object detection capability of PROB the MS-COCO test set. 
    }\vspace{-0.2cm}
    \label{fig:intro}
\end{figure}

Object detection (OD) is a fundamental computer vision task that has a myriad of real-world applications, from autonomous driving~\cite{Roads_OWOD, Driving_OWOD}, robotics \cite{Robots1, Robots2} to healthcare~\cite{HC1, HC2}. 
However, like many other machine learning systems, generalization beyond the training distribution remains challenging~\cite{OpenSetOD} and limits the applicability of existing OD systems.
To facilitate the development of machine learning methods that maintain their robustness in the real world, a new paradigm of learning was developed -- \emph{Open World Learning} (OWL)~\cite{TowardsOWOD, OW-DETR, RevisitingOWOD, UC-OWOD, two_branch_OWOD, OWOD_OCPL,OWEntitySegmentation, OWInstanceSegmentation,OWTracking, owcls1, owcls2}.
In OWL, a machine learning system is tasked with reasoning about both known and unknown  concepts, while slowly learning over time from a non-stationary data stream.
In Open World Object Detection (OWOD), a model is expected to detect all previously learned objects while simultaneously being capable of detecting novel \textit{unknown} objects. These flagged unknown objects can be sent to an oracle (human annotator), which labels the objects of interest. The model is then expected to update itself without catastrophically forgetting previous object classes~\cite{TowardsOWOD}.

While unknown object detection is pivotal to the OWOD objective, existing OWOD methods have very low unknown object recall ($\sim$10\%)~\cite{TowardsOWOD, OW-DETR, RevisitingOWOD, UC-OWOD}. 
As such, it is clear that the field has much to improve to meet its actual goal.
The difficulty of unknown object detection stems from a lack of supervision as, unlike known objects, unknown objects are not labeled. 
Hence, while training OD models, object proposals that include an unknown object would be incorrectly penalized as background.  
Thus far, most OWOD methods have attempted to overcome this challenge by using different heuristics to differentiate between unknown objects and background during training. 
For example, OW-DETR~\cite{OW-DETR} uses a pseudo-labeling scheme where image patches with high backbone feature activation are determined to be unknown objects, and these pseudo-labels are used to supervise the OD model.
In contrast, instead of reasoning about known and unknown objects separately using labels and pseudo-labels, we take a more direct approach. 
We aim to learn a probabilistic model for general ``objectness'' (see Fig.~\ref{fig:intro}).  Any object -- both known and unknown -- should have general features that distinguish them from the background, and the learned objectness can help improve both unknown and known object detection.

Herein, we introduce the Probabilistic Objectness Open World Detection Transformer, \ourMethod. \ourMethod\ incorporates a novel probabilistic objectness head into the standard deformable DETR (D-DETR) model. During training, we alternate between estimating the objectness probability distribution and maximizing the likelihood of known objects. Unlike a classification head, this approach does not require negative examples and therefore does not suffer from the confusion of background and unknown objects. During inference, we use the estimated objectness distribution to estimate the likelihood that each object proposal is indeed an object (see Fig. \ref{fig:intro}).
The resulting model is simple and achieves state-of-the-art open-world performance. 
\textbf{We summarize our contributions as follows:}

\begin{itemize}
\setlength\itemsep{0em}   
\item We introduce \ourMethod\ - a novel OWOD method.  PROB incorporates a probabilistic objectness prediction head that is jointly optimized as a density model of the image features along with the rest of the transformer network. We utilize the objectness head to improve both critical components of OWOD: unknown object detection and incremental learning. 

\item We show extensive experiments on all OWOD benchmarks demonstrating the \ourMethod's capabilities, which outperform all existing OWOD models. On MS-COCO, \ourMethod\ achieves relative gains of 100-300\% in terms of unknown recall over all existing OWOD methods while improving known object detection performance $\sim 10 \%$ across all tasks.

\item We show separate experiments for incremental learning tasks where \ourMethod\ outperformed both OWOD baselines and baseline incremental learning methods.  
\end{itemize}

Code and model of our work will be open-sourced upon publication.

\section{Related Works}
\label{sec:related-work}

\paragraph{Open World Object Detection.} The Open World Object Detection task, recently introduced by Joseph~\etal~\cite{TowardsOWOD}, has already garnered much attention~\cite{OW-DETR,RevisitingOWOD, UC-OWOD, Driving_OWOD, two_branch_OWOD, Roads_OWOD, OWOD_OCPL} due to its possible real-world impact. 
In their work, Joseph~\etal~\cite{TowardsOWOD} introduced ORE, which adapted the faster-RCNN model with feature-space contrastive clustering, an RPN-based unknown detector, and an Energy Based Unknown Identifier (EBUI) for the OWOD objective. 
Yu~\etal~\cite{OWOD_OCPL} attempted to extend ORE by minimizing the overlapping distributions of the known and unknown classes in the embeddings feature-space by setting the number of feature clusters to the number of classes, and showed reduced confusion between known and unknown objects.
Meanwhile, Wu~\etal~\cite{two_branch_OWOD} attempted to extend ORE by introducing a second, localization-based objectness detection head (introduced by Kim~\etal~\cite{NoClass}), and reported gains in unknown object recall, motivating objectness's utility in OWOD. 

Transformer-based methods have recently shown great potential in the OWOD objective when Gupta~\etal~\cite{OW-DETR} adapted the deformable DETR model for the open world objective - and introduced OW-DETR. OW-DETR uses a pseudo-labeling scheme to supervise unknown object detection, where unmatched object proposals with high backbone activation are selected as unknown objects.
Maaz~\etal~\cite{MultimodalT} reported on the high class-agnostic object detection capabilities of Multi-modal Vision Transformers (MViTs). 
They proceeded to utilize MViTs in the supervision of ORE's unknown object detection and reported significant ($\sim 4\times$) gains in its performance. While MViTs do not obey the OWOD objective as they require additional supervision and are trained on much larger datasets with many additional classes, Maaz~\etal's work motivates the possible generalization potential of transformer-based models. 
Recent work in OWOD motivates the use of transformer-based models~\cite{OW-DETR} and the integration of objectness~\cite{two_branch_OWOD} for robust OWOD performance. While previous methods attempted to use objectness estimation~\cite{two_branch_OWOD,OW-DETR}, none directly integrated it into the class prediction itself. 
Unlike previous works, we both introduce a novel method for probabilistically estimating objectness and directly integrate it into the class prediction itself, improving unknown object detection. 

\paragraph{Class Agnostic Object Detection.} Class agnostic object detection (CA-OD) attempts to learn general objectness features given a limited number of labeled object classes. These general features are then used to detect previously unseen object classes. CA-OD methods are expected to localize objects in a class-agnostic fashion. Current SOTA objectness detection~\cite{LDET, NoClass} all address the same issue; datasets are not densely labeled, and therefore one cannot simply decide that a proposed detection is wrong if it does not overlap with any ground truth label. 
Saito~\etal~\cite{LDET} addressed this issue by introducing a custom image augmentation method, BackErase, which pastes annotated objects on an object-free background. 
Kim~\etal~\cite{NoClass} explored the effect of different losses on learning open-world proposals and found that replacing classification with localization losses, which do not penalize false positives, improves performance. 
Unfortunately, the direct integration of CA-OD methods has shown poor OWOD performance. For example, the direct integration of Kim~\etal's~\cite{NoClass} localization-based objectness method into ORE, as presented by Wu~\etal~\cite{two_branch_OWOD}, resulted in a 70\% drop in unknown object recall. Although indirectly, our work integrates insights from CA-OD, e.g., the lack of penalization of false positives.

\section{Background}
\label{sec:background}

\paragraph{Problem Formulation.} 
Let us begin by introducing the notations for standard object detection before extending them to the open-world objective. During training, a model $f$ is trained on a dataset $\mathcal{D} = \{\mathcal{I}, \mathcal{Y}\}$, which contains $K$  known object classes. The dataset contains $N$ images and corresponding labels, $\mathcal{I}=\{\bm{I}_1, \bm{I}_2,\dots,\bm{I}_{N}\} $ and $ \mathcal{Y}=\{\bm{Y}_1, \bm{Y}_2,\dots,\bm{Y}_{N}\}$, respectively. Each label $\bm{Y}_i, i\in[1,2,..., N]$ is composed of $J$ annotated objects $\bm{Y}_i=\{\bm{y}_1, ... \bm{y}_J\}\in \mathcal{Y}$, which is a set of \textit{object} labels, each of which is a vector containing bounding box coordinates and object class label, i.e., $\bm{y}_j = [\bm{l}_j, x_j, y_j, w_j, h_j]$ where $l_j \in  \{0,1\}^{K}$ is a one-hot vector. 

 Let us now extend this formulation to the open-world objective. We follow the formulation introduced by Joseph~\etal~\cite{TowardsOWOD}. Given a task/time $t$, there are $K^t$ known in-distribution classes, and an associated dataset $\mathcal{D}^t = \{\mathcal{I}^t, \mathcal{Y}^t\}$, which contains $N^t$ images and corresponding labels.
 Unlike before, the object class label is now a $K^t+1$ - dimensional vector $l_j \in  \{0,1\}^{K^t+1}$, where the first element is used to represent unknown objects. 
 There may be an unbounded number of unknown classes, but of these, $U^t$ are classes of interest (unknown classes we would like detect). 
 The model then sends the discovered unknown object objects to an oracle (e.g., a human annotator), which will label the new objects of interest. These newly labeled objects are then used to produce $\mathcal{D}^{t+1}$ (which only contains instances of the $U^t$ newly introduced object classes). The model is then updated, given only $\mathcal{D}^{t+1}$, $f^t$, and a limited subset of $\mathcal{D}^{i}, i\in\{0,1,\dots,t\}$ to produce $f^{t+1}$ that can detect $K^{t+1}=K^t+U^t$ object classes. This cycle may be repeated as much as needed. 

\paragraph{DETR for Open World Learning.} 
DETR-type models~\cite{DETR, ddetr} have transformed the object detection field due to their simplified design, which has less inductive bias. These models utilize a transformer encoder-decoder to directly transform spatially-distributed features, encoded using some backbone network, into a set of $N_\text{query}$ object predictions (which can include background predictions). 
The decoder utilizes $N_\text{query}$ learned query vectors, each of which queries the encoded image and outputs a corresponding \textit{query embedding}, $\bm{q} \in \mathbb{R}^D$, i.e., $\bm{Q}=f_\text{feat}^t(I)\in\mathbb{R}^{N_\text{query} \times D}$.
Each query embedding is then input into the bounding box regression ($f_\text{bbox}^t$) and classification ($f_\text{cls}^t$) heads (see Fig.~\ref{fig:method}, bottom). 
The classification head takes each query embedding and predicts whether it belongs to one of the known objects \textit{or} background/unknown object. 
The extension of this formulation to open-world object detection is non-trivial, as the model needs to further separate the background/unknown objects from each other, which is unsupervised (the unknown objects are not labeled). To solve this, OW-DETR~\cite{OW-DETR} incorporated an attention-driven pseudo-labeling scheme where the unmatched queries were scored by the average backbone activation, and the top $u_k$(=5) of them were selected as unknown objects. These pseudo-labels were used during training to supervise unknown object detection, with the inference remaining unchanged.

\section{Method}
\label{sec:method}

\begin{figure*}
    \centering
    \includegraphics[width=0.95\linewidth]{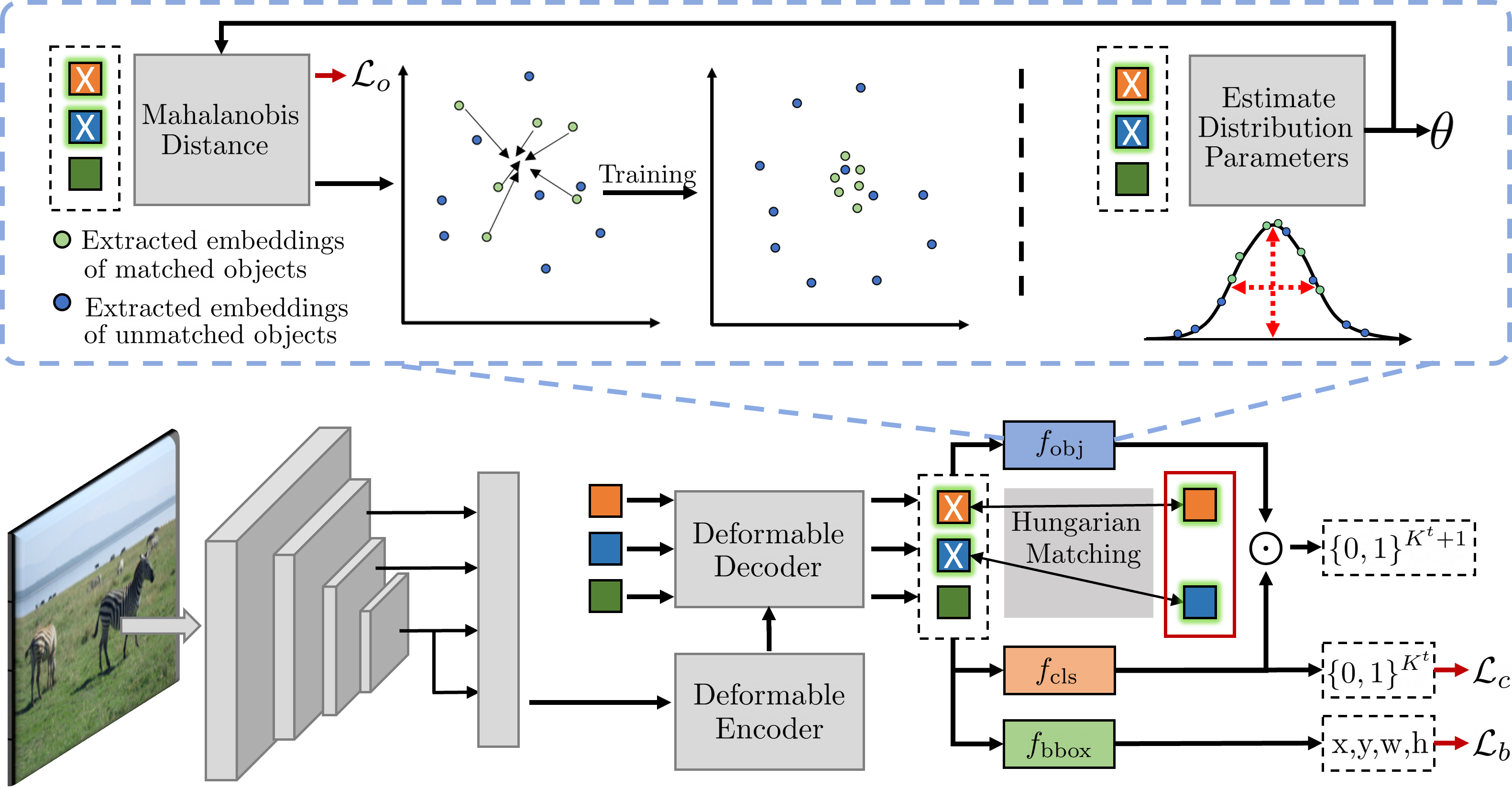}\vspace{-0.2cm}
    \caption{\textbf{Overview of the proposed \ourMethod\ for open-world object detection}. \textbf{(top) Probabilistic objectness module.} 
    The probability distribution parameters of all of the query embeddings, $\theta$, are first estimated via the exponential moving average of the mean and covariance estimators. The Mahalanobis distance is then calculated, and the sum of \textit{matched} query embeddings (green dots in scatterplots and white `X' on query embeddings) are penalized ($\mathcal{L}_o$). This causes the query embeddings of objects to slowly migrate towards the mean, i.e., increased likelihood. 
    \textbf{(bottom) Overview of the entire method.} The base architecture of \ourMethod\ is the deformable DETR (D-DETR) model.
    Query embeddings are produced by the D-DETR model and subsequently used by the classification, bounding box, and objectness heads. The classification head is trained using a sigmoid focal loss ($\mathcal{L}_c$), while the bounding box head is trained with L1 and gIoU losses ($\mathcal{L}_b$). For class prediction, the learned objectness probability multiplies the classification probabilities to produce the final class predictions.
    \vspace{-0.35cm}}
    \label{fig:method}
\end{figure*}

We propose \ourMethod, which adapts the deformable DETR (D-DETR)~\cite{ddetr} model for the open world by incorporating our novel `probabilistic objectness' head. 
In Sec.~\ref{sec:method:2prob_perspective}, we describe how the objectness head is trained and used in inference. In Sec.~\ref{sec:method:5incremental}, we describe how the learned objectness is incorporated in incremental learning; namely, how to learn about new classes in a new task without forgetting old classes. 
Fig.~\ref{fig:method} illustrates the proposed
probabilistic objectness open-world object detection transformer, \ourMethod.

\subsection{Probabilistic Objectness} 
\label{sec:method:2prob_perspective}
The standard D-DETR produces a set of $N_{\text{query}}$ query embedding for every image, each of which is used by the detection heads to produce the final predictions. The extension of D-DETR to the open world objective requires the addition of another class label, ``Unknown Object''. However, unlike the other objects, unknown objects are not labeled -- and therefore, one cannot distinguish between them and background predictions while training. Therefore, most OWOD methods attempt to identify these unknown objects and assign them pseudo-labels during training. 
Rather than directly attempting to identify unknown objects, we propose to separate the object ($o$) and object class ($\bm{l}|o$) predictions. By separately learning about objectness, $p(o|\bm{q})$ and object class probability $p(\bm{l}|o,\bm{q})$, we no longer need to identify unknown objects while training. 
Our modified inference scheme is described as:
\begin{gather}
\label{eq:new_prob}
    p(\bm{l}|\bm{q}) = p(\bm{l}|o,\bm{q})\cdot p(o|\bm{q}).
\end{gather}
The classification head, $f_\text{cls}^t(\bm{q})$ can now operate under the assumption that it already knows if a query embedding represents an object or not, and it learns to imitate $p(\bm{l}|o,\bm{q})$. Meanwhile, our objectness head (introduced below) learns to estimate $ p(o|\bm{q})$. Our final class prediction, in the notations of our modified D-DETR model is: 
\begin{gather}
    p(\bm{l}|\bm{q}) = f_\text{cls}^t(\bm{q})\cdot f_\text{obj}^t(\bm{q}),
\end{gather}
Given a background query, the objectness head should predict a very low probability of it being an object (i.e., $f_\text{obj}^t(\bm{q}) \approx 0$) and suppress the prediction of any objects.  Conversely, if the query contains an object, then the objectness prediction show be high (i.e., $f_\text{obj}^t(\bm{q}) \approx 1$), and the task of classifying the query into any of the known objects or an unknown object is left to the classification head.
Our challenge now becomes learning a good objectness model.

To build a robust objectness model, we turn to probabilistic models. 
We parametrize the objectness probability to be a multivariate Gaussian distribution in the query embedding space, i.e., $o|\bm{q} \sim \mathcal{N}(\bm{\mu}, \bm{\Sigma})$.
To predict objectness, we simply calculate the objectness likelihood, or:
\begin{align} \label{eq:obj_prob}
    f_\text{obj}^t(\bm{q}) &= \exp\big(-(\bm{q}-\bm{\mu})^T \Sigma^{-1} (\bm{q}-\bm{\mu})\big) \\
    &=\exp\big(-d_M(\bm{q})^2\big),
\end{align}
where $d_M$ denotes the Mahalanobis distance for the query embeddings.  
This design choice is well motivated by the current out-of-distribution (OOD) detection literature~\cite{OOD_GMM, E_OOD, FS_OOD}, where class-conditional Gaussian distributions are used to model the feature distribution and detected outliers.  
However, rather than using \textit{class-conditional} Gaussian distributions to model the embedded feature space, we use a \textit{class-agnostic} Gaussian distribution -- as we aim to learn general object features that are shared across all classes.

Training is done in an alternating, two-step process - where we (i) estimate the distribution parameters and (ii) maximize the likelihood of matched embeddings (Fig. \ref{fig:method}). To estimate the query embedding distribution parameters, we estimate the batch- mean, $\bm{\mu}\in \mathbb{R}^D$, and covariance, $\bm{\Sigma}\in \mathbb{R}^{D\times D}$ of the query embeddings $\bm{Q}$ using the empirical mean and covariance estimators with exponential averaging. On the other hand, to maximize the likelihood of matched embeddings, we penalize the squared Mahalanobis distance, $d_M(\bm{q})^2$, of matched query embeddings. 
The objectness loss is, therefore, defined as:
\begin{equation}
    \mathcal{L}_o = \sum_{i\in Z}d_M(\bm{q}_i)^2,
\end{equation}
where Z is a list of indices of the matched queries.

\subsection{Objectness for Incremental Learning}
\label{sec:method:5incremental}
In the OWOD objective, models are expected to incrementally learn newly discovered objects without catastrophically forgetting previously seen objects. To do so, OWOD methods keep a small set of images, or exemplars, to mitigate catastrophic forgetting~\cite{OW-DETR, TowardsOWOD, OWOD_OCPL, RevisitingOWOD, two_branch_OWOD}. While previous methods randomly selected instances/object classes, we believe that actively selecting instances based on their objectness score has the potential to further improve OWOD performance. 
Note, this does not require any extra information since we are only using existing labels, unlike in classic active learning where a model queries an oracle for additional labels. 
Specifically, we select instances that had either low/high objectness as exemplars. 
Instances with low objectness are expected to be relatively difficult instances, as the model was unsure of whether they were an object, and learning them is expected to improve the model performance on newly introduced objects. This is in line with current state-of-the-art active learning methods~\cite{AL_Enthropy}. 
Meanwhile, instances with high objectness are expected to be highly representative of that object class. The selection of these instances is expected to impede catastrophic forgetting, as shown in the incremental learning field~\cite{IL_iCaRL, IL_E2E, IL_3}. Specifically, after training on a particular dataset $\mathcal{D}^t$, we compute the objectness probability of every matched query embedding. We then select the top/bottom 25 scoring objects per object class. In our experiments, to avoid having an unfair advantage, if more images are selected than in previous works~\cite{OW-DETR, TowardsOWOD}, we randomly sub-sample the exemplars to match previous works.

\section{Experiments \& Results}
\begin{table*}[t]
\centering
\caption{\textbf{State-of-the-art comparison for OWOD on M-OWODB (top) and S-OWODB (bottom).} The comparison is shown in terms of unknown class recall (U-Recall) and known class mAP@0.5 (for  previously, currently, and all known objects).
For a fair comparison in the OWOD setting, we compare with the recently introduced ORE~\cite{TowardsOWOD} not employing EBUI (EBUI relies on a held-out set of unknown images, violating the OWOD objective, as shown in~\cite{RevisitingOWOD, OW-DETR}).
\ourMethod\ outperforms all existing OWOD models across all tasks both in terms of U-Recall and known mAP, indicating our models improved unknown and known detection capabilities. 
The smaller drops in mAP between ``Previously known'' and ``Current known'' from the previous task exemplify that the exemplar selection improved our models' incremental learning performance. 
Note that since all 80 classes are known in Task 4, U-Recall is not computed. Only ORE and OW-DETR are compared in S-OWODB, as other methods have not reported results on this benchmark. See Sec.~\ref{sec:res:OWOD_performance} for more details.  \vspace{-0.3cm}}
\setlength{\tabcolsep}{3pt}
\adjustbox{width=\textwidth}{
\begin{tabular}{@{}l|cc|cccc|cccc|ccc@{}}
\toprule
 \textbf{Task IDs} ($\rightarrow$)& \multicolumn{2}{c|}{\textbf{Task 1}} & \multicolumn{4}{c|}{\textbf{Task 2}} & \multicolumn{4}{c|}{\textbf{Task 3}} & \multicolumn{3}{c}{\textbf{Task 4}} \\ \midrule
 
& \cellcolor[HTML]{FFFFED}{U-Recall} & \multicolumn{1}{c|}{\cellcolor[HTML]{EDF6FF}{mAP ($\uparrow$)}} & \cellcolor[HTML]{FFFFED}{U-Recall} & \multicolumn{3}{c|}{\cellcolor[HTML]{EDF6FF}{mAP ($\uparrow$)}} & \cellcolor[HTML]{FFFFED}{U-Recall} & \multicolumn{3}{c|}{\cellcolor[HTML]{EDF6FF}{mAP ($\uparrow$)}} & \multicolumn{3}{c}{\cellcolor[HTML]{EDF6FF}{mAP ($\uparrow$)}}  \\

 & \cellcolor[HTML]{FFFFED}($\uparrow$) & \begin{tabular}[c]{@{}c}Current \\ known\end{tabular} & \cellcolor[HTML]{FFFFED}($\uparrow$) & \begin{tabular}[c]{@{}c@{}}Previously\\  known\end{tabular} & \begin{tabular}[c]{@{}c@{}}Current \\ known\end{tabular} & Both & \cellcolor[HTML]{FFFFED}($\uparrow$) & \begin{tabular}[c]{@{}c@{}}Previously \\ known\end{tabular} & \begin{tabular}[c]{@{}c@{}}Current \\ known\end{tabular} & Both & \begin{tabular}[c]{@{}c@{}}Previously \\ known\end{tabular} & \begin{tabular}[c]{@{}c@{}}Current \\ known\end{tabular} & Both \\ \midrule

ORE*~\cite{TowardsOWOD} & \cellcolor[HTML]{FFFFED} 4.9  & 56.0 & \cellcolor[HTML]{FFFFED}2.9 & 52.7 & 26.0 & 39.4  & \cellcolor[HTML]{FFFFED}3.9 & 38.2 & 12.7 & 29.7 & 29.6 & 12.4 & 25.3 \\ 

UC-OWOD~\cite{UC-OWOD} & \cellcolor[HTML]{FFFFED} 2.4  & 50.7 & \cellcolor[HTML]{FFFFED} 3.4 & 33.1 & 30.5 & 31.8  & \cellcolor[HTML]{FFFFED} 8.7 & 28.8 & 16.3 & 24.6 & 25.6 & 15.9 & 23.2 \\ 

OCPL~\cite{OWOD_OCPL} & \cellcolor[HTML]{FFFFED} 8.26  & 56.6 & \cellcolor[HTML]{FFFFED} 7.65 & 50.6 & 27.5 & 39.1  & \cellcolor[HTML]{FFFFED} 11.9 & 38.7 & 14.7 & 30.7 & 30.7 & 14.4 & 26.7 \\ 

2B-OCD~\cite{two_branch_OWOD} & \cellcolor[HTML]{FFFFED} 12.1  & 56.4 & \cellcolor[HTML]{FFFFED} 9.4 & 51.6 & 25.3 & 38.5  & \cellcolor[HTML]{FFFFED} 11.6 & 37.2 & 13.2 & 29.2 & 30.0 & 13.3 & 25.8 \\ 

OW-DETR~\cite{OW-DETR} & \cellcolor[HTML]{FFFFED}7.5  & 59.2 & \cellcolor[HTML]{FFFFED}6.2 & 53.6 & \textbf{33.5} & 42.9 & \cellcolor[HTML]{FFFFED}5.7 & 38.3 & 15.8 & 30.8 & 31.4 & 17.1 & 27.8 \\

\textbf{Ours: \ourMethod }  & \cellcolor[HTML]{FFFFED} \textbf{19.4}  & \textbf{59.5 }& \cellcolor[HTML]{FFFFED} \textbf{17.4} & \textbf{55.7} & 32.2 & \textbf{44.0}  & \cellcolor[HTML]{FFFFED} \textbf{19.6} & \textbf{43.0} & \textbf{22.2} & \textbf{36.0} & \textbf{35.7} & \textbf{18.9} & \textbf{31.5} \\ 

\midrule
\midrule 

ORE*~\cite{TowardsOWOD} & \cellcolor[HTML]{FFFFED}1.5  & 61.4 & \cellcolor[HTML]{FFFFED}3.9 & 56.5 & 26.1 & 40.6  & \cellcolor[HTML]{FFFFED}3.6 & 38.7 & 23.7 & 33.7 & 33.6 & 26.3 & 31.8 \\

OW-DETR~\cite{OW-DETR} & \cellcolor[HTML]{FFFFED}5.7  & 71.5 & \cellcolor[HTML]{FFFFED}6.2 & 62.8 & 27.5 & 43.8 & \cellcolor[HTML]{FFFFED}6.9 & 45.2 & 24.9 & 38.5 & 38.2 & 28.1 & 33.1 \\


\textbf{Ours: \ourMethod} & \cellcolor[HTML]{FFFFED}\textbf{17.6}  & \textbf{73.4} & \cellcolor[HTML]{FFFFED}\textbf{22.3} & \textbf{66.3} & \textbf{36.0} & \textbf{50.4} & \cellcolor[HTML]{FFFFED}\textbf{24.8} & \textbf{47.8} & \textbf{30.4} & \textbf{42.0} & \textbf{42.6} & \textbf{31.7} & \textbf{39.9} \\
\bottomrule

\end{tabular}%
}\vspace{-0.2cm}
\label{table:t1_owod}
\end{table*}

\begin{table*}[t]

\centering
\caption{\textbf{Impact of progressively integrating our contributions into the baseline.} The comparison is shown in terms of known class average precision (mAP) and unknown class recall (U-Recall) on M-OWODB. All models shown include a finetuning step to mitigate catastrophic forgetting. 
\textbf{\ourMethod}\texttt{-Obj} is our model without objectness likelihood maximization.
\textbf{\ourMethod}\texttt{-IL} is our model without active exemplar selection.
For context, we also include the performance of deformable DETR and an upper bound (D-DETR trained with ground-truth unknown class annotations) as reported by Gupta \etal \cite{OW-DETR}. As all classes are known in Task 4, U-Recall is not computed.
}\vspace{-0.3cm}
\label{tab:ablation_table}
\setlength{\tabcolsep}{2pt}
\adjustbox{width=\textwidth}{
\begin{tabular}{@{}l|cc|cccc|cccc|ccc@{}}
\toprule

 \textbf{Task IDs} ($\rightarrow$)& \multicolumn{2}{c|}{\textbf{Task 1}} & \multicolumn{4}{c|}{\textbf{Task 2}} & \multicolumn{4}{c|}{\textbf{Task 3}} & \multicolumn{3}{c}{\textbf{Task 4}} \\ \midrule
 
 & \cellcolor[HTML]{FFFFED}{U-Recall} & \multicolumn{1}{c|}{\cellcolor[HTML]{EDF6FF}{mAP ($\uparrow$)}} & \cellcolor[HTML]{FFFFED}{U-Recall} & \multicolumn{3}{c|}{\cellcolor[HTML]{EDF6FF}{mAP ($\uparrow$)}} & \cellcolor[HTML]{FFFFED}{U-Recall} & \multicolumn{3}{c|}{\cellcolor[HTML]{EDF6FF}{mAP ($\uparrow$)}} & \multicolumn{3}{c}{\cellcolor[HTML]{EDF6FF}{mAP ($\uparrow$)}}  \\

 & \cellcolor[HTML]{FFFFED}($\uparrow$) & \begin{tabular}[c]{@{}c}Current \\ known\end{tabular} & \cellcolor[HTML]{FFFFED}($\uparrow$) & \begin{tabular}[c]{@{}c@{}}Previously\\  known\end{tabular} & \begin{tabular}[c]{@{}c@{}}Current \\ known\end{tabular} & Both & \cellcolor[HTML]{FFFFED}($\uparrow$) & \begin{tabular}[c]{@{}c@{}}Previously \\ known\end{tabular} & \begin{tabular}[c]{@{}c@{}}Current \\ known\end{tabular} & Both & \begin{tabular}[c]{@{}c@{}}Previously \\ known\end{tabular} & \begin{tabular}[c]{@{}c@{}}Current \\ known\end{tabular} & Both \\ \midrule

Upper Bound & \cellcolor[HTML]{FFFFED}31.6  & 62.5  & \cellcolor[HTML]{FFFFED}40.5 & 55.8 & 38.1 & 46.9 & \cellcolor[HTML]{FFFFED}42.6 & 42.4 & 29.3 & 33.9 & 35.6 & 23.1 & 32.5 \\

\begin{tabular}[c]{@{}l@{}} D-DETR \cite{ddetr} \end{tabular} & -&60.3  & - & 54.5 & 34.4 & 44.7  & - & 40.0 & 17.7 & 33.3 & 32.5 & 20.0 & 29.4 \\ 
\midrule
\midrule

\textbf{\ourMethod}\texttt{-Obj}   & \cellcolor[HTML]{FFFFED} \textbf{21.1} & 39.3 & \cellcolor[HTML]{FFFFED} \textbf{18.9} & 41.0 & 23.5 & 32.3  & \cellcolor[HTML]{FFFFED} \textbf{22.2} & 34.7 & 16.3 & 28.6 & 29.2 & 13.4 & 25.2\\

\textbf{\ourMethod}\texttt{-IL}   & \cellcolor[HTML]{FFFFED} 19.4 & \textbf{59.5} & \cellcolor[HTML]{FFFFED} 15.9 & 54.7 & \textbf{32.2} & 43.5  & \cellcolor[HTML]{FFFFED} 18.4 & 42.6 & 20.7 & 35.3 & 34.7 & 17.4 & 30.4\\

Final: \textbf{\ourMethod }  & \cellcolor[HTML]{FFFFED} 19.4  & \textbf{59.5 }& \cellcolor[HTML]{FFFFED} 17.4 & \textbf{55.7} & \textbf{32.2} & \textbf{44.0}  & \cellcolor[HTML]{FFFFED} 19.6 & \textbf{43.0} & \textbf{22.2} & \textbf{36.0} & \textbf{35.7} & \textbf{18.9} & \textbf{31.5} \\ 
\bottomrule

\end{tabular}%
}\vspace{-0.5em}

\end{table*}

\begin{figure*}
    \includegraphics[width=1\linewidth]{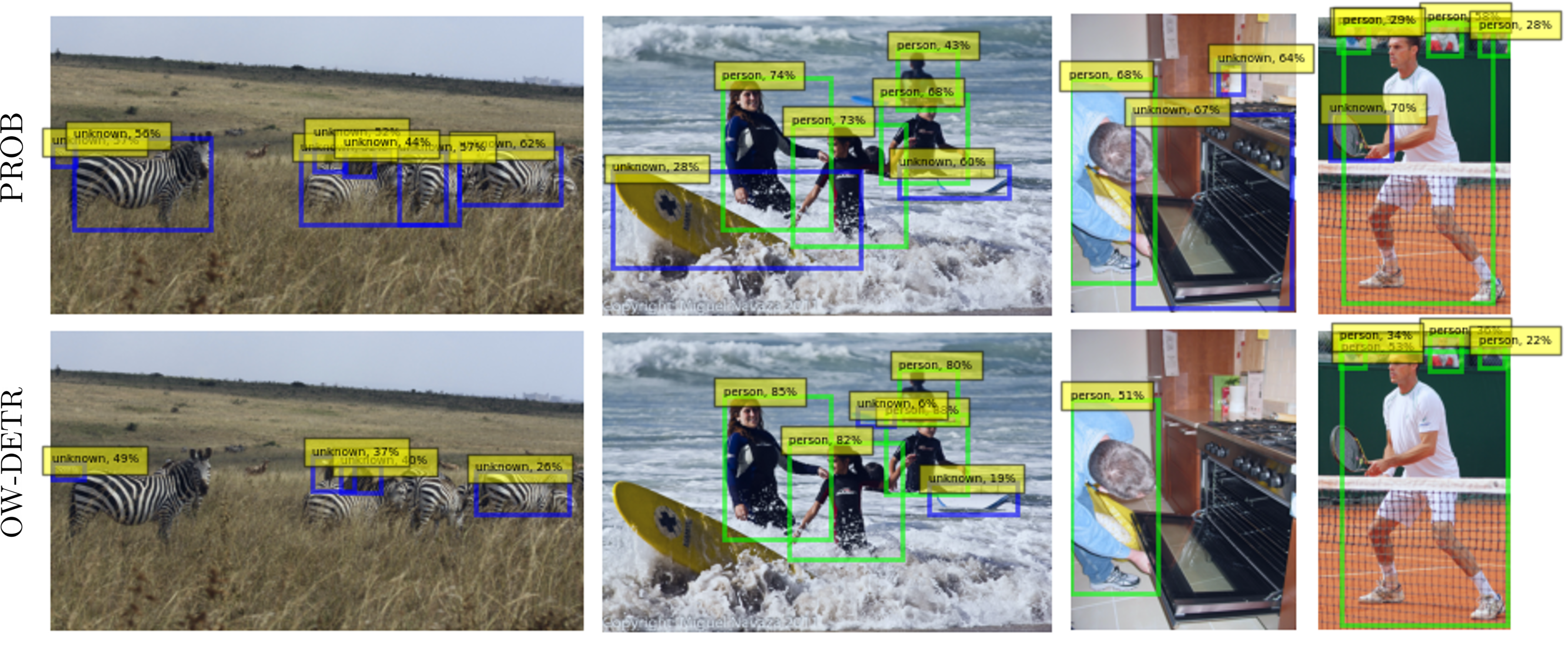}\vspace{-0.2cm}
    \caption{\textbf{Qualitative results on example images from MS-COCO test set}. Detections of \ourMethod\ (top row) and OW-DETR (bottom row) are displayed, with  {\color{green}Green} - known and  {\color{blue}Blue} - unknown object detections. Across all examples, \ourMethod\ detected more unknown objects than OW-DETR, for example, tennis racket in the right column  and zebras in the left column.
    Interestingly, when OW-DETR does detect unknown objects, the predictions have very low confidence, e.g., the surfing board in the center-left column.
    \vspace{-0.35cm}
    }
    \label{fig:qualitative_results}
\end{figure*}

\begin{figure}
    \centering
    \includegraphics[width=1\columnwidth]{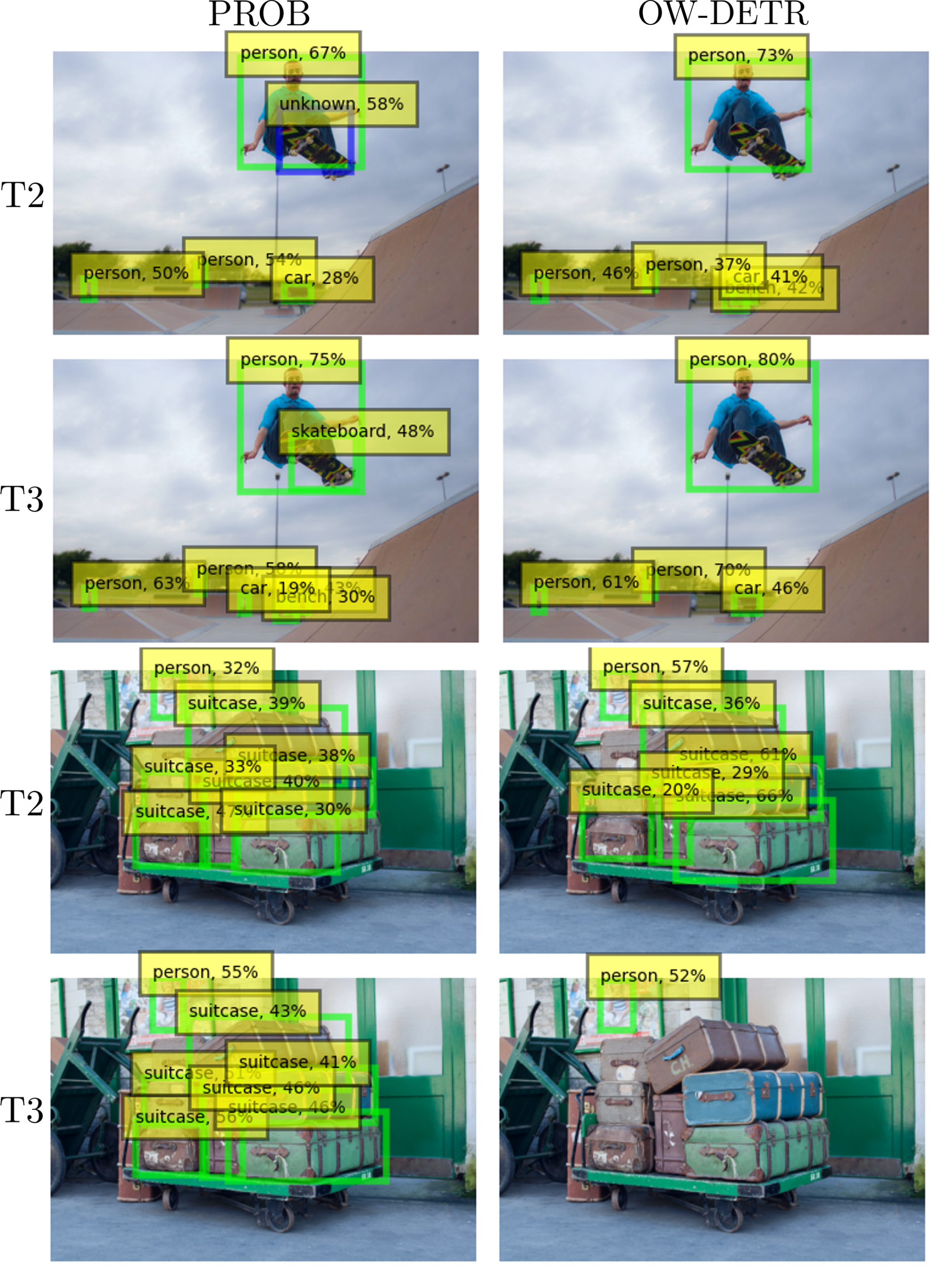}\vspace{-0.2cm}
    \caption{
    \textbf{Qualitative examples of forgetting and improved OWOD}. Detections of \ourMethod\ (left) and OW-DETR (right) are displayed, with  {\color{green}Green} - known and  {\color{blue}Blue} - unknown object detections of the same image across Tasks 2 and 3. (top) \ourMethod\ is able to detect the skateboard as unknown in Task 2, and subsequently classify it in Task 3. (bottom) Example of OW-DETR catastrophically forgetting a previously known object (suitcase).    
    }
    \vspace{-0.3cm}
    \label{fig:more_qual}
\end{figure}

We performed extensive experiments on all OWOD benchmarks, comparing \ourMethod\ to all reported OWOD methods (Sec.~\ref{sec:res:OWOD_performance}). Extensive ablations show the importance of each one of \ourMethod's components while shedding additional light on their function (Sec.~\ref{sec:res:Ablation}). Finally, we test our model's incremental learning performance on the PASCAL VOC 2007 benchmark compared to other OWOD and incremental learning methods (Sec.~\ref{sec:res:iOD}).

\paragraph{Datasets.} We evaluate \ourMethod\  on the benchmarks introduced by Joseph \etal~\cite{TowardsOWOD} and Gupta \etal~\cite{OW-DETR}, which we will reference as ``superclass-mixed OWOD benchmark'' (M-OWODB) and ``superclass-separated OWOD benchmark'' (S-OWODB) respectively. 
Briefly, in M-OWODB, images from MS-COCO~\cite{MSCOCO}, PASCAL VOC2007~\cite{VOC2007}, and PASCAL VOC2012 are grouped into four sets of non-overlapping Tasks; $\{T_1,\cdots,T_4\}$ {s.t.} classes in a task $T_t$ are not introduced until $t$ is reached. 
In each task $T_t$, an additional 20 classes are introduced - and in training for task $t$, only these classes are labeled, while in the test set, all the classes encountered in $\{T_\lambda: \lambda\leq t\}$ need to be detected.
For the construction of S-OWODB, only the MS-COCO dataset was used and a clear separation of super-categories (e.g., animals,  vehicles) was performed. The dataset was split into four Tasks; however, to keep the superclass integrity, a varying number of classes is introduced per increment.
For more, please refer to Joseph~\etal~\cite{TowardsOWOD} and Gupta~\etal~\cite{OW-DETR}, respectively. 

\paragraph{Evaluation Metrics.} For known classes mean average precision (mAP) is used. To better understand the quality of continual learning, mAP is partitioned into previously and newly introduced object classes. 
As common in OWOD, we use unknown object recall (U-recall) as the main metric for unknown object detection~\cite{OW-DETR, OWOD_OCPL, RevisitingOWOD, two_branch_OWOD, MultimodalT} as mAP cannot be used (not all the unknown objects are annotated). 
To study unknown object confusion, we report Absolute Open-Set Error (A-OSE), the absolute number of unknown objects classified as known. For more, see Joseph~\etal~\cite{TowardsOWOD}.\\

\paragraph{Implementation Details.}
We use the deformable DETR~\cite{ddetr} model utilizing multi-scale features extracted via a DINO-pretrained~\cite{DINO} Resnet-50 FPN backbone~\cite{OW-DETR}. The deformable transformer then extracts $N_{\text{query}}=100$ and $D=256$ dimensional query embeddings as discussed above. The embedding probability distribution is estimated by calculating the exponential moving average of the mean and covariance of the query embeddings over the mini-batches (with a batch size of 5), with a momentum of 0.1. Additional details are provided in the appendix. 

\subsection{Open World Object Detection Performance}
\label{sec:res:OWOD_performance}
\ourMethod's OWOD performance, compared with all other reported OWOD methods on their respective benchmarks, can be seen in Tab.~\ref{table:t1_owod}. While all methods reported results on M-OWODB, OWOD performance on the recently introduced S-OWODB is only reported by OW-DETR. S-OWODB is expected to be a more difficult benchmark for unknown object detection, as there is complete super-category separation across the Tasks (i.e., it is more difficult to generalize from animals to vehicles than from dogs to cats). \ourMethod\ shows substantial improvement in unknown object recall (U-Recall), with additional improvements in known object mAP compared to all other OWOD methods.    

\paragraph{Unknown Object Detection.} Across all four Tasks and both benchmarks, \ourMethod's unknown object detection capability, quantified by U-Recall, is 2-3x of those reported in previous state-of-the-art OWOD methods. This result exemplifies the utility of the proposed probabilistic objectness formulation in the OWOD objective. Other OWOD methods have attempted to integrate objectness, most notably OW-DETR~\cite{OW-DETR} with their class-agnostic classification head, and 2B-OCD~\cite{two_branch_OWOD}, with its localization-based objectness head (which was first reported by Kim~\etal). As reported by Kim~\etal~\cite{NoClass}, indeed, the utilization of their localization-based objectness estimation improves unknown object recall, as can be seen by the $\sim4$-point improvement between 2B-OCD and OW-DETR. However, \ourMethod\ outperformed both methods in terms of U-Recall and known mAP across all Tasks. This shows the relative robustness of our probabilistic framework compared to other methods that incorporated objectness for improved unknown object detection in the OWOD setting. When looking at Sup. Tab.~\ref{tab:wi_ose}, it becomes evident that \ourMethod\ not only detects more unknowns (as exemplified by the increased U-Recall), but it also does so much more accurately, as can be seen by the reduction in A-OSE. For example, \ourMethod\ had an A-OSE of 5195, 6452, and 2641 to OW-DETR's 10240, 8441, 6803 for Tasks 1-3, respectively.

\paragraph{Known Object Detection and Incremental Learning.} \ourMethod\ progressively outperforms all previous stat-of-the-art OWOD methods in terms of known object mAP. Compared to OW-DETR, the method with the closest performance to ours, \ourMethod\ increased mAP 0.3, 1.1, 5.2, and 3.7 known object mAP on Tasks 1-4, M-OWODB (Tab.~\ref{table:t1_owod}, top), and 1.9, 6.6, 3.5, and 6.8 known object on Tasks 1-4, S-OWODB  (Tab.~\ref{table:t1_owod}, bottom). The improvement in mAP, even in Task 1, suggests that the learned objectness also improved OWOD known object handling. The relatively smaller drops between `previously known' and the previous task's `both' (e.g., on M-OWODB, between Tasks 1-2 OW-DETR's mAP dropped 5.6 while \ourMethod\ dropped 3.8, even though it started with higher overall mAP).

\paragraph{Qualitative Results.}
Fig.~\ref{fig:qualitative_results} shows qualitative results on example images from MS-COCO. The detections for known  (green) and unknown (blue) objects are shown for \ourMethod\ and OW-DETR.
We observe that \ourMethod\ has better unknown object performance (e.g., zebras in the left image). The unknown object prediction themselves are much more confident cup and surfing board in the two center images).
In Fig.~\ref{fig:more_qual}, \ourMethod\ detected the skateboard in Task 2 and subsequently learned it in Task 3, while OW-DETR missed both. \ourMethod\ is also less prone to catastrophically forgetting an entire object class, as can be seen on the bottom of Fig.~\ref{fig:more_qual}, where OW-DETR catastrophically forgot `suitcase' in between Task 2 and 3, while \ourMethod\ did not.
These results exemplify that \ourMethod\ has promising OWOD performance. Additional qualitative results are presented in the appendix.

\subsection{Ablation Study}
\label{sec:res:Ablation}
Tab. \ref{tab:ablation_table} shows results from our ablation study. 
\textbf{\ourMethod} \texttt{- Obj} disables the objectness likelihood maximization step during training, and now the objectness head only estimates the embedding probability distribution. As can be seen in Tab.~\ref{tab:ablation_table}, this had the effect of slightly increasing the unknown recall but drastically reducing the known object mAP across all Tasks. While counterintuitive, this sheds some light on how our method actually functions. 
Without maximizing the likelihood of the matched query embeddings, the objectness prediction becomes random. As it no longer suppresses background query embeddings, the model then predicts a lot of background patches as unknown 
objects and objects -- known and unknown -- as background. 
As a result, the known class mAP drops because some known object predictions are suppressed by the random objectness prediction. \textbf{\ourMethod} \texttt{- IL} disables the active exemplar selection, and it shows the  advantage of using the probabilistic objectness for exemplar selection. Interestingly, it seems that the active selection mostly benefits unknown object recall, with less significant gains in both previously and currently introduced objects for Tasks 2-4. For Task 1, both methods are the same, as no exemplar replay is used. 
In Tab.~\ref{tab:ablation_table}, we also included the reported performance of two D-DETR models: a model trained on all classes (i.e., both known and unknown) denoted as the ``Upper Bound'', and the D-DETR model trained and evaluated only on the known classes.
Comparing the known object mAP of the oracle and D-DETR, it suggests that learning about unknowns, rather than just treating them the same way as background, leads to improved known object detection capabilities. This possibly explains why \ourMethod\ had a higher known mAP compared to OW-DETR in Task 1 on both benchmarks. 

\begin{table}
\centering
\caption{\textbf{State-of-the-art comparison for incremental object detection (iOD) on PASCAL VOC.} 
The comparison is shown in terms of new, old, and overall mAP. In each setting, the model is first trained on $10$, $15$ or $19$ classes, and then the additional $10$, $5$, and $1$ class(es) are introduced. \ourMethod\ achieves favorable performance in comparison to existing approaches in all three settings. See Sec.~\ref{sec:res:iOD} for additional details.\vspace{-0.3cm}}
\label{tab:siOD}
\setlength{\tabcolsep}{8pt}
\adjustbox{width=0.47\textwidth}{%
\begin{tabular}{@{}lccc@{}}
\toprule
{\color[HTML]{009901} \textbf{10 + 10 setting}} & old classes & new classes & final mAP\\ \midrule
ILOD \cite{ILOD} & 63.2  & 63.2 & 63.2 \\

Faster ILOD \cite{F_ILOD} & 69.8  & 54.5 & 62.1 \\ 

ORE $-$ EBUI~\cite{TowardsOWOD} & 60.4 & 68.8 & 64.5 \\ 

OW-DETR\cite{OW-DETR} & 63.5  &  67.9 & 65.7 \\ 
\midrule

\textbf{Ours: \ourMethod} & 66.0 & 67.2 & \textbf{66.5} \\

\midrule\midrule

{\color[HTML]{009901} \textbf{15 + 5 setting}} & old classes & new classes & final mAP \\ \midrule
ILOD \cite{ILOD} & 68.3 &  58.4 & 65.8 \\

Faster ILOD \cite{F_ILOD} & 71.6 & 56.9 & 67.9 \\ 

ORE $-$ EBUI~\cite{TowardsOWOD} & 71.8 & 58.7 & 68.5 \\ 
OW-DETR \cite{OW-DETR}& 72.2 & 59.8 & 69.4 \\ 
\midrule
\textbf{Ours: \ourMethod} & 73.2 & 60.8 & \textbf{70.1} \\ 
\midrule\midrule

{\color[HTML]{009901} \textbf{19 + 1 setting}} &old classes & new classe & final mAP\\ \midrule
ILOD \cite{ILOD} & 68.5  & 62.7 & 68.2 \\

Faster ILOD \cite{F_ILOD} & 68.9 & 61.1 & 68.5 \\

ORE $-$ EBUI~\cite{TowardsOWOD} & 69.4 & 60.1 & 68.8 \\ 
OW-DETR \cite{OW-DETR} & 70.2 & 62.0 & 70.2 \\
\midrule
\textbf{Ours: \ourMethod} & 73.9  &  48.5 &   \textbf{ 72.6 } \\ 

\bottomrule
\end{tabular}
}

\end{table}

\subsection{Incremental Object Detection}
\label{sec:res:iOD}
As reported by Gupta \etal~\cite{OW-DETR}, the detection of unknowns does seem to improve the incremental learning capabilities of object detection models. This, combined with the introduced improved exemplar selection, results in \ourMethod\ performing favorably on the incremental object detection (iOD) task. 
Tab.~\ref{tab:siOD} shows a comparison of  \ourMethod\ with existing methods on PASCAL VOC 2007, with evaluations performed as reported in~\cite{OW-DETR, TowardsOWOD}. In each evaluation, the model is first trained on 10/15/19 object classes, and then an additional 10/5/1 classes are incrementally introduced. Our model had a final mAP of 66.5, 70.1, and 72.6 to OW-DETR's 65.7, 69.4, and 70.2, respectively. Results with class-breakdown can be found in the appendix.


\section{Conclusions}
The Open World Object Detection task is a complex and multifaceted objective, integrating aspects of generalized open-set object detection and incremental learning. 
For robust OWOD methods to function, understanding and detecting the unknown is critical. 
We proposed a novel probabilistic objectness-based approach to tackle the OWOD objective, which significantly improves this critical aspect of the benchmark. The proposed \ourMethod\ integrates the introduced probabilistic objectness into the deformable DETR model, adapting it to the open-world setting. 
Our ablations shed light on the inner workings of our method while motivating the use of each one of its components.
Extensive experiments show that \ourMethod\ significantly outperforms all existing OWOD methods on all OWOD benchmarks. However, much room for improvement remains, both in unknown object detection and other aspects of the OWOD task. As such, probabilistic models may have great potential for the OWOD objective, creating powerful algorithms that can reliably operate in the open world.

{\small
\bibliographystyle{ieee_fullname}
\bibliography{egbib}
}

\clearpage
\appendix

\begin{minipage}[t]{\columnwidth}
\centering
\Large
\bf
Supplementary Materials
\end{minipage}

\begin{figure}[!h]
    \centering
    \includegraphics[width=1\columnwidth]{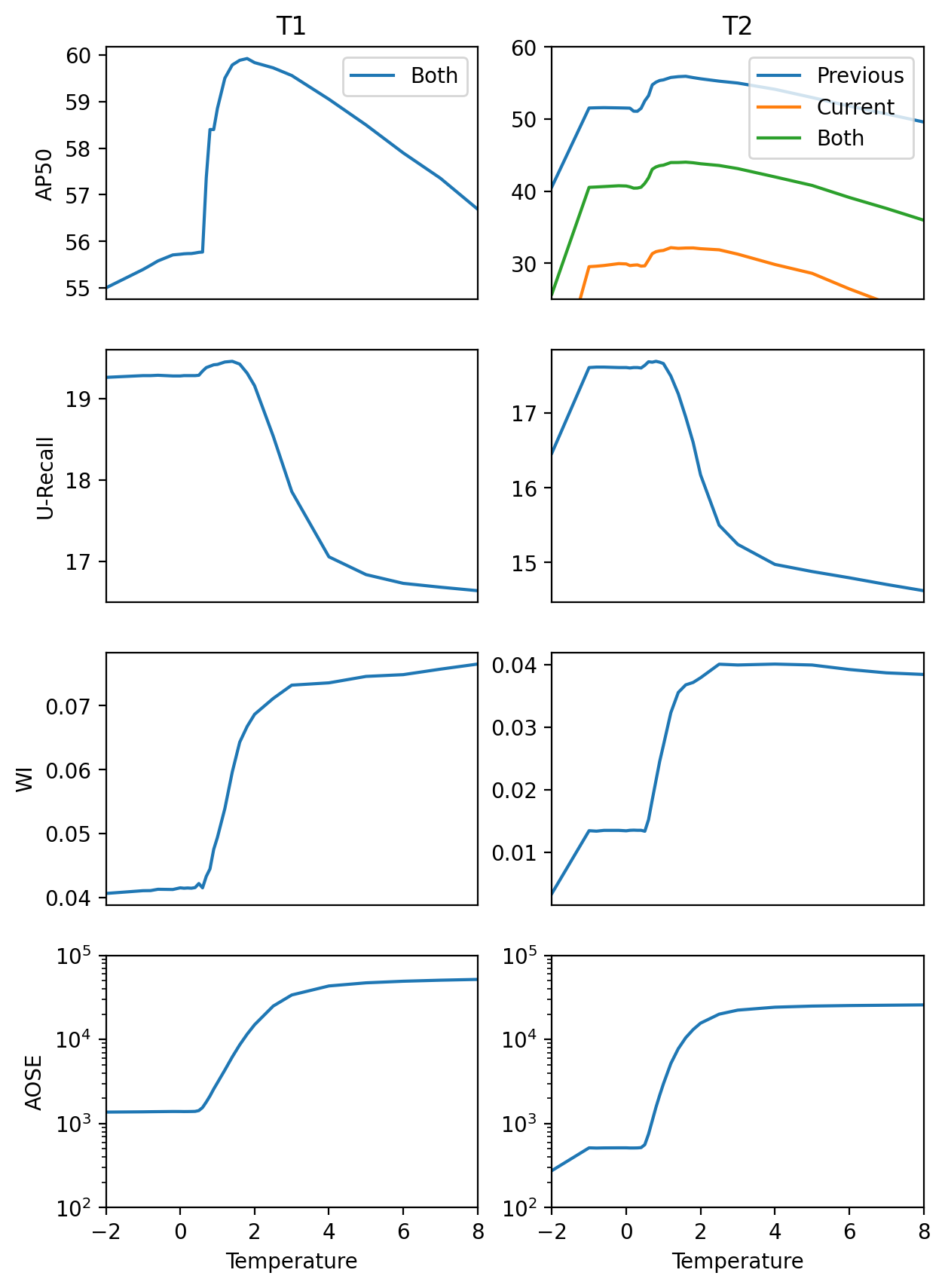}\vspace{-0.2cm}
    \caption{\textbf{Objectness Temperature Sweep Experiment}. Effect of objectness temperature on model performance for Task 1 and Task 2. WI and A-OSE monotonically increase with objectness temperature, while U-Recall and AP50 peak at an objectness temperature of 1.0-1.6. See Sec.~\ref{sec:temperature_effect} for additional details.
    }
    \label{fig:sup_temp_sweep}
\end{figure}

\section{Additional Quantitative Results}
\label{sec:sup_quant}
\subsection{Evaluation using WI and A-OSE Metrics}
\label{sec:sup:wi_aose}
In Sec.~\ref{sec:res:OWOD_performance}, we referred to additional metrics that quantify unknown-known object confusion. 
Tab.~\ref{tab:wi_ose} shows a comparison of the different open-world object detection (OWOD) methods on the M-OWODB dataset~\cite{TowardsOWOD} in terms of unknown recall (U-recall), wilderness impact (WI), and absolute open-set error (A-OSE). 
U-Recall measures the models' ability to detect unknown objects and indicates the degree of unknown objects that are detected by an OWOD method. 
Meanwhile, the Wilderness Impact (WI) and Absolute Open Set Error (A-OSE) measure the model's confusion in predicting an unknown instance as a known class.
Specifically, WI measures the effect the unknown detections have on the model's precision. WI is problematic as U-Recall grows, unknown object precision becomes more dominant, causing WI to increase - even if the models have the same unknown object precision. Therefore, the WI will typically increase with U-Recall. Meanwhile, A-OSE  measures the total number of unknown instances detected as one of the known classes and is less affected by U-Recall.

When examining Tab.~\ref{tab:wi_ose}, it becomes apparent that \ourMethod\ outperforms all other OWOD methods in terms of U-Recall while having lower (or similar) A-OSE. While the A-OSE reduction in Tab.~\ref{tab:wi_ose} is not very large, reducing the objectness temperature (see Sec.~\ref{sec:temperature_effect}) results in a much lower A-OSE. 
For example, for Task 2, at an objectness temperature of 0.5, \ourMethod's OWOD performance can be seen in Tab.~\ref{tab:wi_ose}. Specifically, A-OSE drops to 562, with only marginal degradation of the known mAP, as can be seen in Fig.~\ref{fig:sup_temp_sweep}. WI additionally reduces by almost 50\% to $0.0133$. 
This shows that our model can be easily tuned towards either better unknown or known precision simply by changing the objectness temperature.

\begin{table*}[t]
\centering
\caption{ \textbf{Uknown Object Confusion on M-OWODB.} The comparison is shown in terms of wilderness impact (WI), absolute open set error (A-OSE) and unknown class recall (U-Recall). The unknown recall (U-Recall) metric quantifies a model's ability to retrieve the unknown object instances. 
\ourMethod\ achieves improved WI, A-OSE and U-Recall over OW-DETR across tasks, thereby indicating less confusion in detecting unknown instances as known classes with higher unknown instance detection capabilities. See Sec.~\ref{sec:sup:wi_aose} for additional details.\vspace{-0.3cm}}
\label{tab:wi_ose}
\setlength{\tabcolsep}{10pt}
\adjustbox{width=\textwidth}{
\begin{tabular}{@{}l|ccc|ccc|ccc}
\toprule
 \textbf{Task IDs} ($\rightarrow$)& \multicolumn{3}{c|}{\textbf{Task 1}} & \multicolumn{3}{c|}{\textbf{Task 2}} & \multicolumn{3}{c}{\textbf{Task 3}} \\
\midrule

 & \cellcolor[HTML]{FFFFED}{U-Recall} & \cellcolor[HTML]{EDF6FF}{WI} & \cellcolor[HTML]{EDF6FF}{A-OSE} & \cellcolor[HTML]{FFFFED}{U-Recall} & \cellcolor[HTML]{EDF6FF}{WI} & \cellcolor[HTML]{EDF6FF}{A-OSE}  & \cellcolor[HTML]{FFFFED}{U-Recall} & \cellcolor[HTML]{EDF6FF}{WI} & \cellcolor[HTML]{EDF6FF}{A-OSE} \\

 & \cellcolor[HTML]{FFFFED}($\uparrow$) & \cellcolor[HTML]{EDF6FF}($\downarrow$) & \cellcolor[HTML]{EDF6FF}($\downarrow$) & \cellcolor[HTML]{FFFFED}($\uparrow$) & \cellcolor[HTML]{EDF6FF}($\downarrow$) & \cellcolor[HTML]{EDF6FF}($\downarrow$) & \cellcolor[HTML]{FFFFED}($\uparrow$) & \cellcolor[HTML]{EDF6FF}($\downarrow$) & \cellcolor[HTML]{EDF6FF}($\downarrow$) \\

 \midrule



ORE $-$ EBUI ~\cite{TowardsOWOD} & \cellcolor[HTML]{FFFFED}4.9  & 0.0621 & 10459 & \cellcolor[HTML]{FFFFED}2.9 & 0.0282 & 10445 & \cellcolor[HTML]{FFFFED}3.9 & 0.0211 & 7990  \\

2B-OCD ~\cite{two_branch_OWOD} & \cellcolor[HTML]{FFFFED}12.1 & 0.0481 & - & \cellcolor[HTML]{FFFFED}9.4 & 0.160 & - & \cellcolor[HTML]{FFFFED}11.6 & 0.0137 & -  \\

OW-DETR \cite{OW-DETR} & \cellcolor[HTML]{FFFFED}7.5  & 0.0571 & 10240 & \cellcolor[HTML]{FFFFED}6.2 & 0.0278 & 8441 & \cellcolor[HTML]{FFFFED}5.7 & 0.0156 & 6803  \\

OCPL ~\cite{OWOD_OCPL} & \cellcolor[HTML]{FFFFED} 8.3  & \textbf{0.0413} & 5670 & \cellcolor[HTML]{FFFFED} 7.6 & 0.0220 & 5690 & \cellcolor[HTML]{FFFFED}11.9 & 0.0162 & 5166  \\
\midrule
\midrule

\textbf{Ours: \ourMethod} & \cellcolor[HTML]{FFFFED} \textbf{19.4}  & 0.0569 & 5195 & \cellcolor[HTML]{FFFFED} 17.4 & 0.0344 & 6452 & \cellcolor[HTML]{FFFFED} 19.6 & 0.0151 &2641  \\

\textbf{Ours: \ourMethod$(\tau=0.5)$} & \cellcolor[HTML]{FFFFED} 19.3  & 0.0415 & \textbf{1428} & \cellcolor[HTML]{FFFFED} \textbf{17.7} & \textbf{0.0133} & \textbf{562 }& \cellcolor[HTML]{FFFFED} \textbf{19.7} & \textbf{0.0082} & \textbf{387}  \\
\bottomrule

\end{tabular}%
}

\end{table*}

\begin{table*}[t]
\centering
\caption{\textbf{State-of-the-art comparison for incremental object detection (iOD) on PASCAL VOC.} We experiment on 3 different settings. The comparison is shown in terms of per-class AP and overall mAP. The $10$, $5$ and $1$ class(es) in \colorbox{gray}{gray} background are introduced to a detector trained on the remaining $10$, $15$ and $19$ classes, respectively. \ourMethod\ achieves favorable performance in comparison to existing OWOD approaches in all three settings. See Sec.~\ref{sec:sup:iOD} for additional details.\vspace{-0.3cm}}
\label{tab:iOD}
\setlength{\tabcolsep}{3pt}
\adjustbox{width=\textwidth}{%
\begin{tabular}{@{}lccccccccccccccccccccc@{}}
\toprule
{\color[HTML]{009901} \textbf{10 + 10 setting}} & aero & cycle & bird & boat & bottle & bus & car & cat & chair & cow & table & dog & horse & bike & person & plant & sheep & sofa & train & tv & mAP \\ \midrule
ILOD \cite{ILOD} & 69.9 & 70.4 & 69.4 & 54.3 & 48 & 68.7 & 78.9 & 68.4 & 45.5 & 58.1 & \cellcolor[HTML]{EEEEEE}59.7 & \cellcolor[HTML]{EEEEEE}72.7 & \cellcolor[HTML]{EEEEEE}73.5 & \cellcolor[HTML]{EEEEEE}73.2 & \cellcolor[HTML]{EEEEEE}66.3 & \cellcolor[HTML]{EEEEEE}29.5 & \cellcolor[HTML]{EEEEEE}63.4 & \cellcolor[HTML]{EEEEEE}61.6 & \cellcolor[HTML]{EEEEEE}69.3 & \cellcolor[HTML]{EEEEEE}62.2 & 63.2 \\

Faster ILOD \cite{F_ILOD} & 72.8 & 75.7 & 71.2 & 60.5 & 61.7 & 70.4 & 83.3 & 76.6 & 53.1 & 72.3 & \cellcolor[HTML]{EEEEEE}36.7 & \cellcolor[HTML]{EEEEEE}70.9 & \cellcolor[HTML]{EEEEEE}66.8 & \cellcolor[HTML]{EEEEEE}67.6 & \cellcolor[HTML]{EEEEEE}66.1 & \cellcolor[HTML]{EEEEEE}24.7 & \cellcolor[HTML]{EEEEEE}63.1 & \cellcolor[HTML]{EEEEEE}48.1 & \cellcolor[HTML]{EEEEEE}57.1 & \cellcolor[HTML]{EEEEEE}43.6 & 62.1 \\ 
ORE $-$ (CC  + EBUI)~\cite{TowardsOWOD} & 53.3 & 69.2 & 62.4 & 51.8 & 52.9 & 73.6 & 83.7 & 71.7 & 42.8 & 66.8 & \cellcolor[HTML]{EEEEEE}46.8 & \cellcolor[HTML]{EEEEEE}59.9 & \cellcolor[HTML]{EEEEEE}65.5 & \cellcolor[HTML]{EEEEEE}66.1 & \cellcolor[HTML]{EEEEEE}68.6 & \cellcolor[HTML]{EEEEEE}29.8 & \cellcolor[HTML]{EEEEEE}55.1 & \cellcolor[HTML]{EEEEEE}51.6 & \cellcolor[HTML]{EEEEEE}65.3 & \cellcolor[HTML]{EEEEEE}51.5 & 59.4 \\
ORE $-$ EBUI~\cite{TowardsOWOD} & 63.5 & 70.9 & 58.9 & 42.9 & 34.1 & 76.2 & 80.7 & 76.3 & 34.1 & 66.1 & \cellcolor[HTML]{EEEEEE}56.1 & \cellcolor[HTML]{EEEEEE}70.4 & \cellcolor[HTML]{EEEEEE}80.2 & \cellcolor[HTML]{EEEEEE}72.3 & \cellcolor[HTML]{EEEEEE}81.8 & \cellcolor[HTML]{EEEEEE}42.7 & \cellcolor[HTML]{EEEEEE}71.6 & \cellcolor[HTML]{EEEEEE}68.1 & \cellcolor[HTML]{EEEEEE}77 & \cellcolor[HTML]{EEEEEE}67.7 & 64.5 \\ 
OW-DETR\cite{OW-DETR} & 61.8 & 69.1 & 67.8 & 45.8 & 47.3 & 78.3 & 78.4 & 78.6 & 36.2 & 71.5 &  \cellcolor[HTML]{EEEEEE} 57.5 &  \cellcolor[HTML]{EEEEEE} 75.3 &  \cellcolor[HTML]{EEEEEE} 76.2 &  \cellcolor[HTML]{EEEEEE} 77.4 &  \cellcolor[HTML]{EEEEEE} 79.5 &  \cellcolor[HTML]{EEEEEE} 40.1 &  \cellcolor[HTML]{EEEEEE} 66.8 &  \cellcolor[HTML]{EEEEEE} 66.3 &  \cellcolor[HTML]{EEEEEE} 75.6 & \cellcolor[HTML]{EEEEEE} 64.1 & 65.7 \\ 
\midrule

\textbf{Ours: \ourMethod} & 70.4 & 75.4 & 67.3 & 48.1 & 55.9 & 73.5 & 78.5 & 75.4 & 42.8 & 72.2&  \cellcolor[HTML]{EEEEEE} 64.2 &  \cellcolor[HTML]{EEEEEE} 73.8 &  \cellcolor[HTML]{EEEEEE} 76.0 &  \cellcolor[HTML]{EEEEEE} 74.8 &  \cellcolor[HTML]{EEEEEE} 75.3 &  \cellcolor[HTML]{EEEEEE} 40.2 &  \cellcolor[HTML]{EEEEEE} 66.2 &  \cellcolor[HTML]{EEEEEE} 73.3 &  \cellcolor[HTML]{EEEEEE} 64.4 & \cellcolor[HTML]{EEEEEE} 64.0 & \textbf{66.5} \\

\midrule\midrule

{\color[HTML]{009901} \textbf{15 + 5 setting}} & aero & cycle & bird & boat & bottle & bus & car & cat & chair & cow & table & dog & horse & bike & person & plant & sheep & sofa & train & tv & mAP \\ \midrule
ILOD \cite{ILOD} & 70.5 & 79.2 & 68.8 & 59.1 & 53.2 & 75.4 & 79.4 & 78.8 & 46.6 & 59.4 & 59 & 75.8 & 71.8 & 78.6 & 69.6 & \cellcolor[HTML]{EEEEEE}33.7 & \cellcolor[HTML]{EEEEEE}61.5 & \cellcolor[HTML]{EEEEEE}63.1 & \cellcolor[HTML]{EEEEEE}71.7 & \cellcolor[HTML]{EEEEEE}62.2 & 65.8 \\

Faster ILOD \cite{F_ILOD} & 66.5 & 78.1 & 71.8 & 54.6 & 61.4 & 68.4 & 82.6 & 82.7 & 52.1 & 74.3 & 63.1 & 78.6 & 80.5 & 78.4 & 80.4 & \cellcolor[HTML]{EEEEEE}36.7 & \cellcolor[HTML]{EEEEEE}61.7 & \cellcolor[HTML]{EEEEEE}59.3 & \cellcolor[HTML]{EEEEEE}67.9 & \cellcolor[HTML]{EEEEEE}59.1 & 67.9 \\ 
ORE $-$ (CC  + EBUI)~\cite{TowardsOWOD} & 65.1 & 74.6 & 57.9 & 39.5 & 36.7 & 75.1 & 80 & 73.3 & 37.1 & 69.8 & 48.8 & 69 & 77.5 & 72.8 & 76.5 & \cellcolor[HTML]{EEEEEE}34.4 & \cellcolor[HTML]{EEEEEE}62.6 & \cellcolor[HTML]{EEEEEE}56.5 & \cellcolor[HTML]{EEEEEE}80.3 & \cellcolor[HTML]{EEEEEE}65.7 & 62.6 \\
ORE $-$ EBUI~\cite{TowardsOWOD} & 75.4 & 81 & 67.1 & 51.9 & 55.7 & 77.2 & 85.6 & 81.7 & 46.1 & 76.2 & 55.4 & 76.7 & 86.2 & 78.5 & 82.1 & \cellcolor[HTML]{EEEEEE}32.8 & \cellcolor[HTML]{EEEEEE}63.6 & \cellcolor[HTML]{EEEEEE}54.7 & \cellcolor[HTML]{EEEEEE}77.7 & \cellcolor[HTML]{EEEEEE}64.6 & 68.5 \\ 
OW-DETR \cite{OW-DETR}& 77.1 & 76.5 & 69.2 & 51.3 & 61.3 & 79.8 & 84.2 & 81.0 & 49.7 & 79.6 & 58.1 & 79.0 & 83.1 & 67.8 & 85.4 & \cellcolor[HTML]{EEEEEE}33.2 & \cellcolor[HTML]{EEEEEE}65.1 & \cellcolor[HTML]{EEEEEE}62.0 & \cellcolor[HTML]{EEEEEE}73.9 & \cellcolor[HTML]{EEEEEE}65.0 & 69.4 \\ 
\midrule
\textbf{Ours: \ourMethod} & 77.9 &77.0 &77.5 &56.7 &63.9 &75.0 &85.5 &82.3 &50.0 &78.5 &63.1 &75.8 &80.0 &78.3 &77.2 &  \cellcolor[HTML]{EEEEEE} 38.4 &  \cellcolor[HTML]{EEEEEE} 69.8 &  \cellcolor[HTML]{EEEEEE} 57.1 &  \cellcolor[HTML]{EEEEEE} 73.7 & \cellcolor[HTML]{EEEEEE} 64.9 & \textbf{70.1} \\ 
\midrule\midrule

{\color[HTML]{009901} \textbf{19 + 1 setting}} & aero & cycle & bird & boat & bottle & bus & car & cat & chair & cow & table & dog & horse & bike & person & plant & sheep & sofa & train & tv & mAP \\ \midrule
ILOD \cite{ILOD} & 69.4 & 79.3 & 69.5 & 57.4 & 45.4 & 78.4 & 79.1 & 80.5 & 45.7 & 76.3 & 64.8 & 77.2 & 80.8 & 77.5 & 70.1 & 42.3 & 67.5 & 64.4 & 76.7 & \cellcolor[HTML]{EEEEEE}62.7 & 68.2 \\

Faster ILOD \cite{F_ILOD} & 64.2 & 74.7 & 73.2 & 55.5 & 53.7 & 70.8 & 82.9 & 82.6 & 51.6 & 79.7 & 58.7 & 78.8 & 81.8 & 75.3 & 77.4 & 43.1 & 73.8 & 61.7 & 69.8 & \cellcolor[HTML]{EEEEEE}61.1 & 68.5 \\
ORE $-$ (CC  + EBUI)~\cite{TowardsOWOD} & 60.7 & 78.6 & 61.8 & 45 & 43.2 & 75.1 & 82.5 & 75.5 & 42.4 & 75.1 & 56.7 & 72.9 & 80.8 & 75.4 & 77.7 & 37.8 & 72.3 & 64.5 & 70.7 & \cellcolor[HTML]{EEEEEE}49.9 & 64.9 \\
ORE $-$ EBUI~\cite{TowardsOWOD} & 67.3 & 76.8 & 60 & 48.4 & 58.8 & 81.1 & 86.5 & 75.8 & 41.5 & 79.6 & 54.6 & 72.8 & 85.9 & 81.7 & 82.4 & 44.8 & 75.8 & 68.2 & 75.7 & \cellcolor[HTML]{EEEEEE}60.1 & 68.8 \\ 
OW-DETR \cite{OW-DETR} & 70.5 & 77.2 & 73.8 & 54.0 & 55.6 & 79.0 & 80.8 & 80.6 & 43.2 & 80.4 & 53.5 & 77.5 & 89.5 & 82.0 & 74.7 & 43.3 & 71.9 & 66.6 & 79.4 & \cellcolor[HTML]{EEEEEE}62.0 & 70.2 \\
\midrule
\textbf{Ours: \ourMethod} & 80.3 &78.9 &77.6 &59.7 &63.7 &75.2 &86.0 &83.9 &53.7 &82.8 &66.5 &82.7 &80.6 &83.8 &77.9 &48.9 &74.5 &69.9 &77.6  &\cellcolor[HTML]{EEEEEE} 48.5 &   \textbf{ 72.6 } \\ 

\bottomrule
\end{tabular}%
}\vspace{-0.8em}

\end{table*}

\begin{figure*}
\centering
    \includegraphics[width=0.97\linewidth]{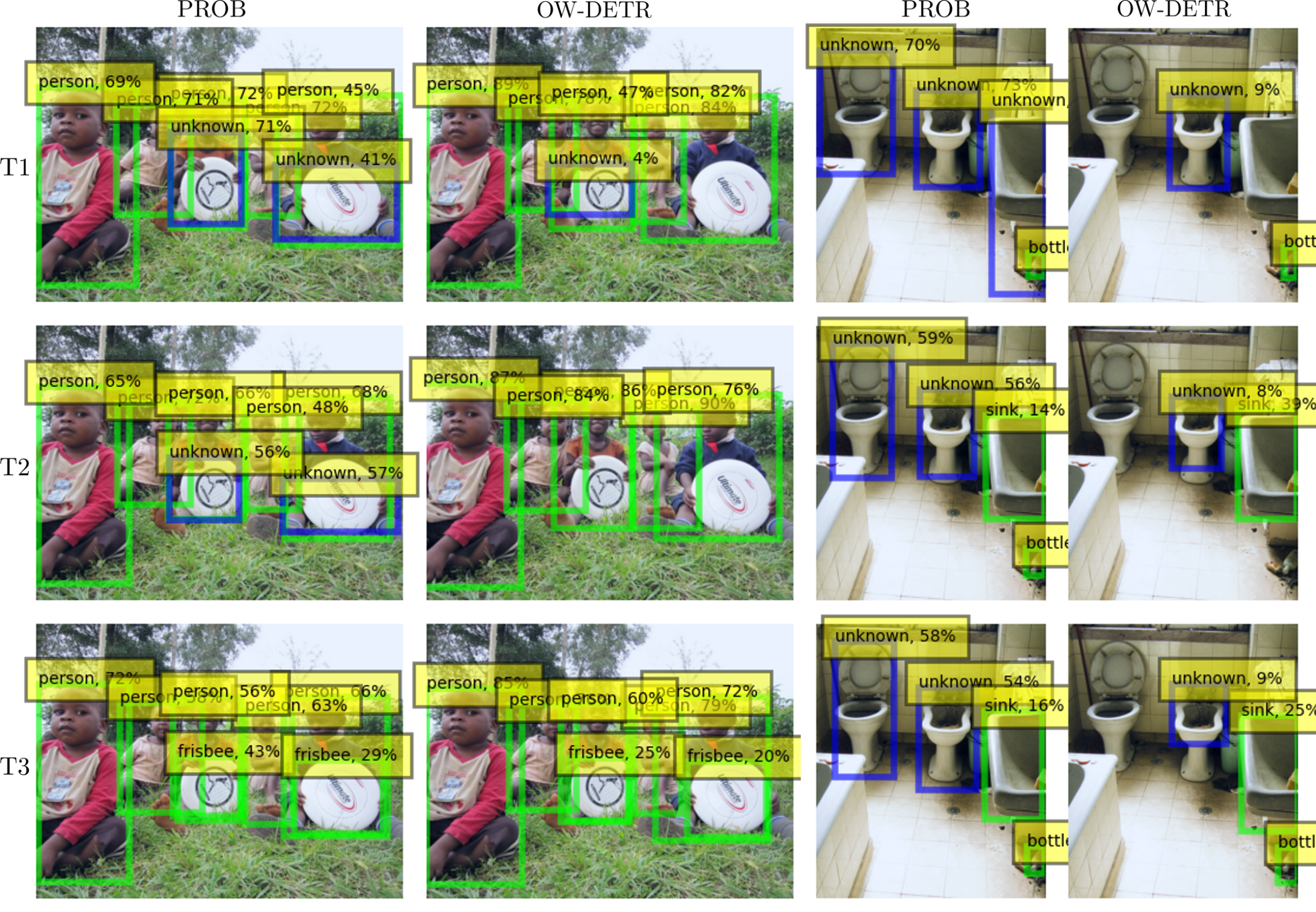}\vspace{-0.2cm}
    \caption{\textbf{General OWOD performance.} \ourMethod\ appears to have favorable OWOD performance, detecting unknowns and incrementally learning them, while OW-DETR does not. For example, in the right column, 
    \ourMethod\ detected both toilets across all tasks and also detected the sink in Task 1 before learning it in Task 2, while OW-DETR did not. When OW-DETR did detect an unknown object, it tended to have very low confidence ($\sim 10\%$), while \ourMethod\ had confidence on par with that of the known objects, showing that \ourMethod\ had more balanced unknown-known predictions. OW-DETR additionally `forgot' the bottle in Task 2. 
    }\vspace{-0.2cm}
    \label{fig:sup_better_owod}
\end{figure*}

\subsection{Incremental Learning}
\label{sec:sup:iOD}
In Sec.~\ref{sec:res:iOD}, we reported \ourMethod's incremental learning capabilities. For completeness, we add the per-class AP of the incremental learning experiments reported in Tab.~\ref{tab:siOD} in Tab.~\ref{tab:iOD}. 
As discussed in Sec. \ref{sec:method:5incremental}, OWOD methods rely on exemplar replay for mitigating catastrophic forgetting. First introduced by Prabhu~\etal~\cite{ExemplarSelectionIL}, exemplar replay adds an additional fine-tuning stage on a balanced set of exemplars after every training step and was shown to be extremely effective. 
Nonetheless, \ourMethod's \textit{active} selection of high/low objectness scoring exemplars further improved both unknown and known object detection performance, as shown in our ablations (see Tab. \ref{tab:ablation_table}). 
Interestingly, the known object performance seems to grow over time (with a delta of 0.5, 0.7, and 0.9 in Tasks 2, 3, and 4, respectively), further motivating the utility of \ourMethod's active exemplar selection.

\subsection{Objectness Temperature Variation}
\label{sec:temperature_effect}
In Sec.~\ref{sec:method:2prob_perspective}, Eq.~\ref{eq:obj_prob}, we introduced the proposed objectness prediction. However, there is an additional hyperparameter, specifically `objectness temperature', which can be varied (see Sec.~\ref{sec:imp_detail}). Objectness temperature does not affect training and can be varied at test time if needed. It controls the degree of confidence in objectness prediction, with higher objectness temperature resulting in less confident objectness prediction. In all our experiments, the objectness temperature was set to 1.3. 
To investigate its effect on model inference, we evaluated the model at different objectness temperatures (see Fig.~\ref{fig:sup_temp_sweep}).
Fig.~\ref{fig:sup_temp_sweep} shows that U-Recall and known mAP have optimum  points; however, the two do not necessarily coincide, as can be seen in Task 2. Meanwhile, there is a definite trade-off between the models' unknown-known object confusion (as quantified by WI and A-OSE) and U-Recall and mAP. A-OSE gets as low as 1300 in Task 1 and 500 in Task 2 (excluding the drop-off at ($\tau=-2$), with the know mAP remaining reasonably high at $55.8$ (dropping from $59.5$) and $40.7$ (dropping from $44.0$). The same can be said of the WI. It is worth noting that, for the methods that have lower A-OSE and WI in Tab.~\ref{tab:wi_ose}, there exists an objectness temperature where \ourMethod\ outperforms them in terms of known mAPs, U-Recall, WI, and A-OSE.

\section{Additional Qualitative Results}
\label{sec:sup_qual}
In Sec.~\ref{sec:res:OWOD_performance}, we noted that  
\ourMethod\ has favorable qualitative performance on the M-OWODB. Specifically, \ourMethod\ consistently detects unknown objects across Tasks, and learns them properly when the objects are incrementally learned. For example, in Fig.~\ref{fig:sup_better_owod}, left, \ourMethod\ detects both frisbees in Task 1 and 2 as `unknown objects' and in Task 3, after frisbees are added as a known class, as frisbees. Meanwhile, OW-DETR detected only one frisbee in Task 1, none in Task 2, and both as frisbees in Task 3. In Fig.~\ref{fig:sup_better_owod}, right, \ourMethod\ detected both toilets and the sink as `unknown' while OW-DETR detected only one of the toilets across all tasks. OW-DETR additionally `forgets' the bottle in Task 2, and remembers it again in Task 3, while \ourMethod\ consistently detected the bottle. When examining the confidence scores of known and unknown objects, it is clear that \ourMethod\ makes balanced predictions of known and unknown objects (i.e., predictions with similar confidence scores), while OW-DETR predicts unknown objects with very low confidence. This imbalance may lead to difficulty in detecting unknown objects at test time, as unknown proposals may not meet the detection confidence threshold.

\begin{figure*}
\centering
    \includegraphics[width=0.955\linewidth]{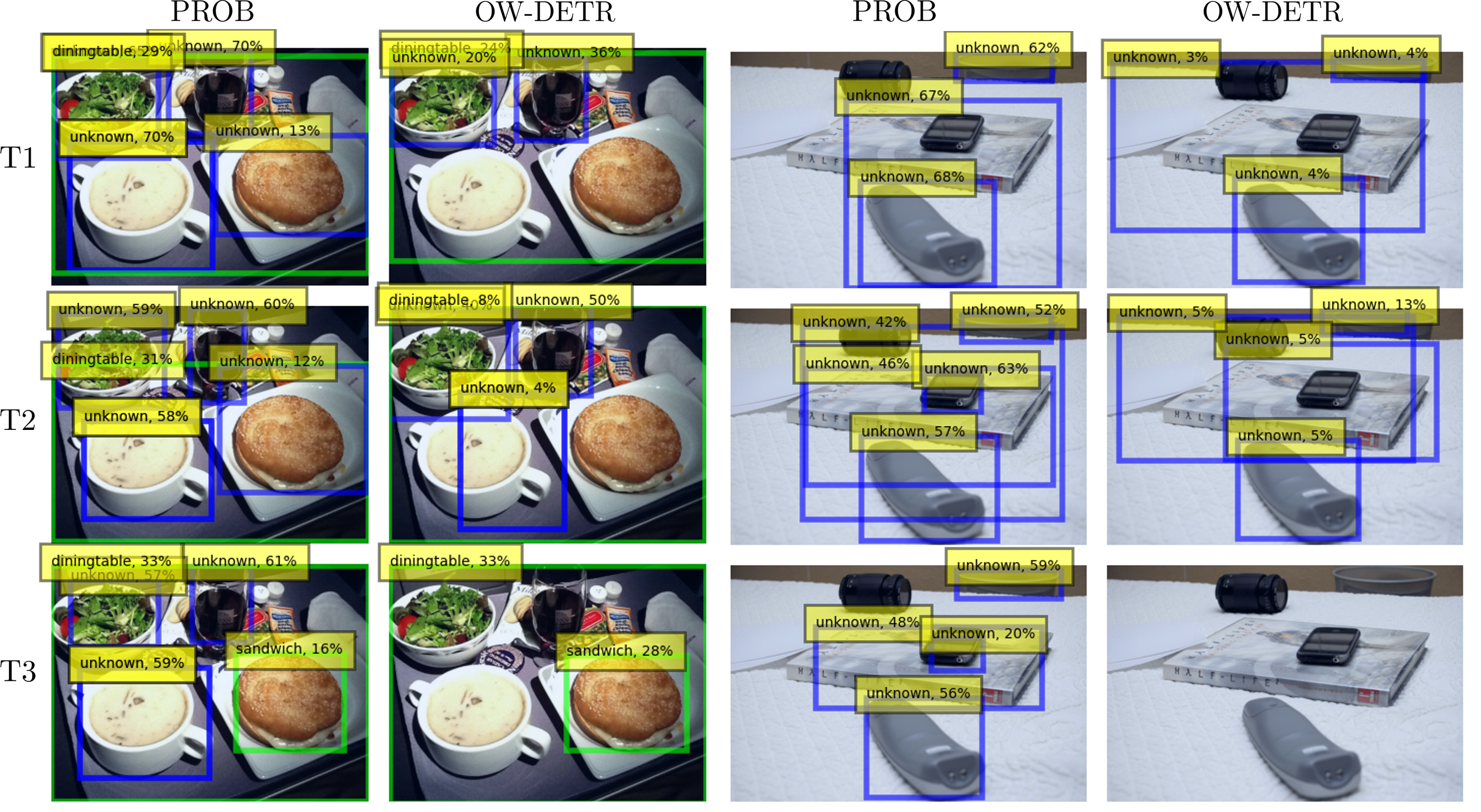}\vspace{-0.2cm}
    \caption{\textbf{Unknown object catastrophic forgetting.} \ourMethod\ does not catastrophically forget unknown objects, while OW-DETR does. 
    For example, OW-DETR catastrophically forgot all previously detected unknown objects when learning for Task 3, while \ourMethod\ did not. 
   After learning for Task 3, OW-DETR no longer detects the bowls (which it previously detected) while \ourMethod\ continued to detect them.}
   \label{fig:sup_unk_forget}
\end{figure*}

\paragraph{Unknown Object Confidence and Forgetting.} 
\ourMethod\ detects unknown objects more confidently and does not forget the unknown objects in later tasks. For example, in Fig.~\ref{fig:sup_unk_forget}, most of the unknown object detections of OW-DETR have less than $<10\%$ confidence. 
Unlike OW-DETR's unknown detection performance, which seems to degrade/fluctuate over time, \ourMethod's unknown object detection seems to improve. 
For example, In Fig.~\ref{fig:sup_unk_forget}, left, \ourMethod\ consistently detected the four unknown objects in the image, with little the bounding box localization and confidence variation across tasks, while OW-DETR's bound box localization, confidence, and final predictions seem erratic, with it not detecting any unknown objects in T3. In Fig.~\ref{fig:sup_unk_forget}, right, there is another example of OW-DETR catastrophically forgetting all unknown objects in the image. Here, OW-DETR detected the remote, book, and trash can in Tasks 1 and 2, but forgot all of them in Task 3, unlike \ourMethod.

\begin{figure*}
\centering
    \includegraphics[width=0.955\linewidth]{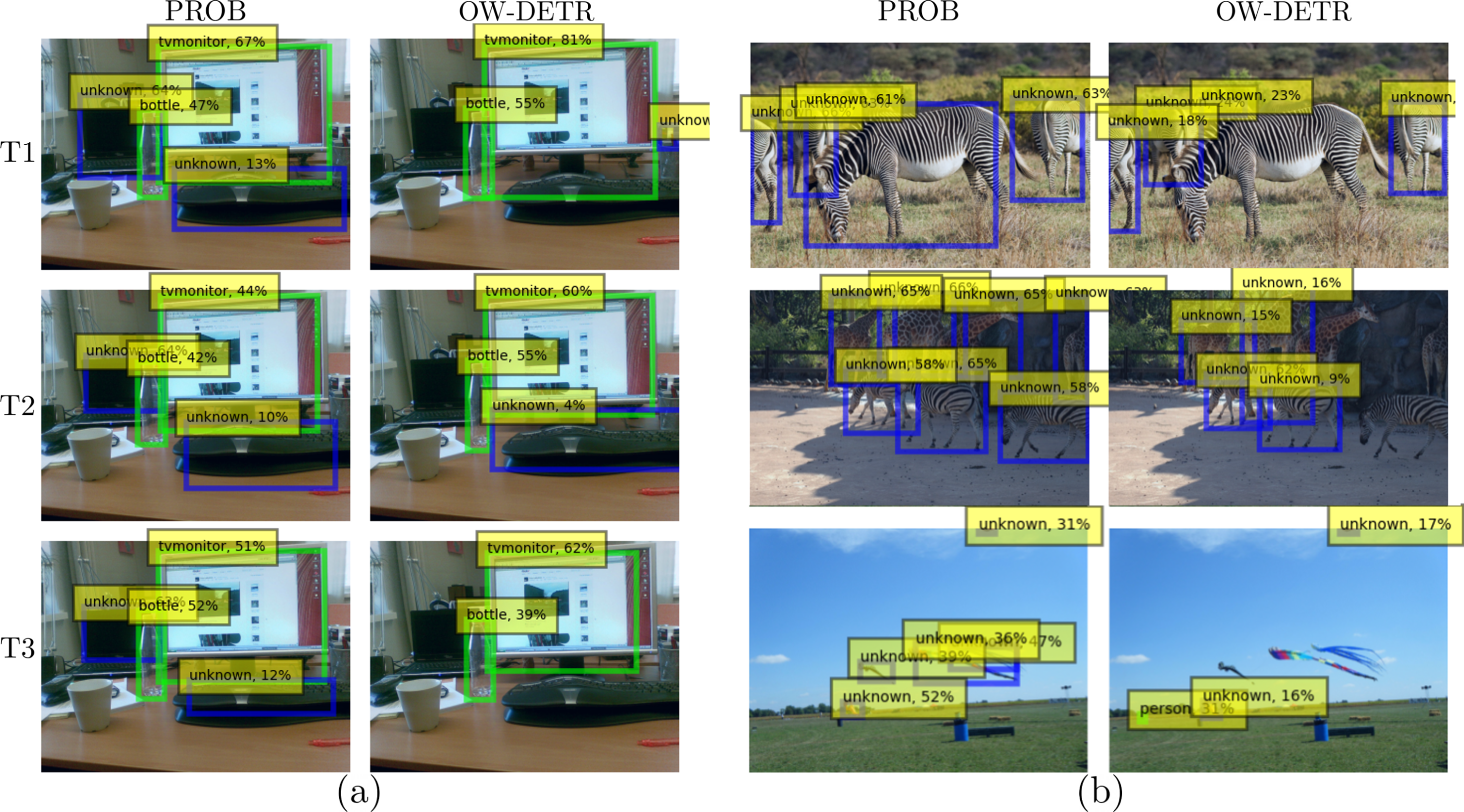}\vspace{-0.2cm}
    \caption{\textbf{Uknown object detection consistency on M-OWODB.} (a) depicts the predictions of \ourMethod\ and OW-DETR on the same image across different tasks. \ourMethod\ consistently detects the same unknown objects as such, while OW-DETR does not. (b) examples of \ourMethod\ and OW-DETR predictions. Even in the same image, OW-DETR does not detect obvious examples of unknown objects (e.g., foreground zebra in the top image), while \ourMethod\ consistently detects unknown objects from the same class.}
    \label{fig:sup_stable}
\end{figure*}

\paragraph{Unknown Object Detection Consistency.} 
Qualitatively \ourMethod\ tends to detect unknown objects more consistently, both across tasks (Fig.~\ref{fig:sup_stable} (a)) and across unknown object instances (Fig.~\ref{fig:sup_stable} (b)). 
In Fig.~\ref{fig:sup_stable} (a), you can see that \ourMethod\ consistently detects the same objects across tasks (e.g., laptop and keyboard) while OW-DETR detects a different unknown object in every Task. 
In Fig.~\ref{fig:sup_stable} (b), you can see that \ourMethod\ tends to detect all the unknown objects of the same class (e.g., zebras, giraffes, and kites), while OW-DETR seems to miss obvious instances of the same object class. 
For example, in Fig.~\ref{fig:sup_stable} (b), top, OW-DETR misses the zebra in the foreground, seemingly only detecting the zebras more in the background. 
This shows the relatively poor performance of OW-DETR in detecting unknown objects, as it doesn't seem to generalize unknown objects across the same class. This puts into question what object features OW-DETR extracts to make its predictions, as they do not seem robust, unlike \ourMethod\, which seems to do this quite well.

\section{Additional Implementation Details}
\label{sec:imp_detail}
We found that assuming that the channels are iid distributed did not result in a change in performance while reducing training time and improving model stability, i.e., $$\bm{\Sigma}=\bm{I}\cdot\bm{\sigma}.$$
This iid assumption makes inverting $\bm{\Sigma}$ trivial and easily computed during training with little effect on training time. When training with an unrestricted $\bm{\Sigma}$, heavy regularization of the matrix inversion calculation of the covariance, required for stable training, causes it to become diagonal. 
For objectness prediction (inference/evaluation \textbf{only}), we added an objectness temperature hyperparameter, $\tau$, $$f_\text{obj}^t(\bm{q}) =\exp\big(-\tau \cdot d_M(\bm{q})^2\big),$$
which was set to 1.3 in all of our experiments ($\tau=1.3$), based on our temperature sweep experiments (see Sec.~\ref{sec:temperature_effect}). 
\ourMethod\ is then trained end-to-end with the joint loss: $$\mathcal{L}=\mathcal{L}_\text{c}+\mathcal{L}_\text{b}+\alpha\mathcal{L}_\text{o},$$
with four Nvidia A100 40GB GPUs, with a batch size of 5. 
The learning rate was taken to be $2\times10^{-3}$, $\beta_1=0.9$, $\beta_2=0.999$, weight decay of $10^{-4}$, and a learning rate drop after 35 epochs by a factor of 10. For finetuning during the incremental learning step, the learning rate is reduced by a factor of 10. All other hyperparameters were taken as reported in OW-DETR \cite{OW-DETR}.

\section{Limitations and Social Impacts}
While the OWOD field has been rapidly progressing, much improvement is required to reach the more nuanced aspects of the OWOD objective. 
As known and unknown object detection rely on different forms of supervision, their predictions are imbalanced with respect to the relation of prediction score and confidence. Exploration of energy-based models \cite{ClassIsEBM} could be a solution to this problem while enabling better separation of known and unknown object classes in the embedded feature space. 
Additional work is still needed in better benchmark design. Current benchmarks expose an entire dataset of novel objects per task. As unknown object recall improves, OWOD algorithms should begin attempting to only discover unknown objects detected by the model. This essentially adds an additional active learning stage between incremental learning steps.

Open-world learning bridges the gap between benchmarks and the real world. In doing so, OWOD algorithms will encounter situations with social impact. To 
detect new objects, OWOD relay on unknown object detection, which may be biased given the initial training dataset. Future research should not only look at unknown object detection capabilities but also its possible biases. To do so, it would be useful to break down unknown object detection capabilities into the relevant subclasses. 
Future models should integrate `forgetting' capabilities that can be applied to particular object classes out of legal and/or privacy concerns. Finally, saving actual images as exemplars may also constitute privacy violations in the open world. As OWOD methods are deployed in the real world, images selected as exemplars will inevitably contain not only the known but also other unknown object classes. These images will be stored as part of the algorithm's lifetime and may contain private or sensitive information. Future work should work on either replacing or censoring selected exemplars to avoid such situations.


 \vfill\null

\end{document}